\documentclass[preprint]{myElsarticle}

\usepackage[total={450pt, 595pt}]{geometry}
\usepackage{setspace}
\usepackage{etoolbox}
\usepackage{graphicx}
\usepackage{amsmath}
\usepackage{amsthm}
\usepackage{amssymb}
\usepackage{amsfonts}
\usepackage[caption=false]{subfig}
\usepackage{threeparttable}
\usepackage{color}
\usepackage{array}
\usepackage{booktabs}
\usepackage{mathrsfs}
\usepackage{multirow}
\usepackage{url}
\usepackage{hyperref}
\usepackage{microtype}

\newtheorem{theorem}{Theorem}
\newtheorem{lemma}{Lemma}
\newdefinition{definition}{Definition}
\newdefinition{assumption}{Assumption}
\newdefinition{condition}{Condition}
\newdefinition{remark}{Remark}
\newdefinition{example}{Example}

\AtBeginEnvironment{theorem}{\doublespacing}
\AtBeginEnvironment{lemma}{\doublespacing}
\AtBeginEnvironment{proof}{\doublespacing}
\AtBeginEnvironment{definition}{\doublespacing}
\AtBeginEnvironment{assumption}{\doublespacing}
\AtBeginEnvironment{condition}{\doublespacing}
\AtBeginEnvironment{remark}{\doublespacing}
\AtBeginEnvironment{example}{\doublespacing}

\newcolumntype{C}[1]{>{\centering\let\newline\\\arraybackslash\hspace{0pt}}m{#1}}
\newcolumntype{L}{>{\centering\arraybackslash}m{4cm}}

\begin{document}

\title{Appropriateness of Performance Indices for Imbalanced Data Classification: An Analysis}

\author[1]{Sankha Subhra Mullick}
\ead{sankha\_r@isical.ac.in}

\author[2]{Shounak Datta}
\ead{shounak.jaduniv@gmail.com}

\author[3]{Sourish Gunesh Dhekane}
\ead{sourishdhekane@gmail.com}

\author[1]{Swagatam Das\corref{cor1}}
\ead{swagatam.das@isical.ac.in}


\cortext[cor1]{Corresponding author}

\address[1]{Electronics and Communication Sciences Unit, Indian Statistical Institute, Kolkata, India}
\address[2]{Department of Electrical and Computer Engineering, Duke University, Durham, NC, USA}
\address[3]{Department of Computer Science and Engineering, Indian Institute of Information Technology, Guwahati, India}

\begin{abstract}
Indices quantifying the performance of classifiers under class-imbalance, often suffer from distortions depending on the constitution of the test set or the class-specific classification accuracy, creating difficulties in assessing the merit of the classifier. We identify two fundamental conditions that a performance index must satisfy to be respectively resilient to altering number of testing instances from each class and the number of classes in the test set. In light of these conditions, under the effect of class imbalance, we theoretically analyze four indices commonly used for evaluating binary classifiers and five popular indices for multi-class classifiers. For indices violating any of the conditions, we also suggest remedial modification and normalization. We further investigate the capability of the indices to retain information about the classification performance over all the classes, even when the classifier exhibits extreme performance on some classes. Simulation studies are performed on high dimensional deep representations of subset of the ImageNet dataset using four state-of-the-art classifiers tailored for handling class imbalance. Finally, based on our theoretical findings and empirical evidence, we recommend the appropriate indices that should be used to evaluate the performance of classifiers in presence of class-imbalance.
\end{abstract}


\maketitle

\section{Introduction}

\subsection{Overview}
Classification is a fundamental supervised learning problem where the task is to develop classifiers (for example, $k$-Nearest Neighbor ($k$NN) \cite{bishop2006pattern}, Multi-Layer Perceptron (MLP) \cite{rumelhart1986}, Support Vector Machine (SVM) \cite{bishop2006pattern} etc.) which can approximate a many-to-one mapping from a set $X$ of $d$-dimensional data points to a set $\mathcal{C}=\{1, 2, \cdots C\}$ of class labels. An allied challenge comes in the form of designing indices \cite{Sokolova2006BeyondAF} which can accurately evaluate the performance of a classifier, considering the particular nature of the classification problem as well as the pertinent data irregularities \cite{das2018}. 

Class imbalance \cite{he2013, guo2017} is a form of data irregularity which is fairly common in many real-world classification problems \cite{guo2017, branco2016, das2018} such as like medical diagnosis, fraud detection, etc. A training set $P \subseteq X$ is considered as class imbalanced when it does not contain equal number of training instances from all the classes (especially those corresponding to the rare and therefore important events). This leads the classifier to be biased in favor of the majority classes and consequently suffers from higher misclassification on the minority classes. Evidently, such bias should be properly compensated during performance evaluation. This indicates the need for special indices, unlike the widely used Accuracy measure which lays more stress on the performance over the majority classes, being unsuitable in presence of class imbalance \cite{japkowicz2006question}.

Over the years, for a binary imbalanced classification problem indices like Recall, Specificity, and Precision \cite{buckland1994} were considered to be the basic measures of performance. 
However, by design, Recall (or Sensitivity) measures the accuracy over the minority (positive) class, Specificity does the same for the majority (negative) class, and Precision \cite{buckland1994} considers the fraction of positives which are accurately classified (true positive) to the number of instances predicted as positives. In other words, these three measures offer different criteria of evaluation by respectively focusing on true positive, true negative (analogous to true positive for the negative class), and false positive (negative instances wrongly classified as positive) counts, and a good classifier is expected to optimize all of them. However, optimizing multiple indices simultaneously is difficult in practice, especially if a trade-off is required. Therefore, attempts were made to combine two or more of these basic indices together to form new measures which can consider multiple distinct aspects during evaluation as well as provide easy interpretability. For example, the GMean \cite{kubat1997} index is calculated by taking the geometric mean of Sensitivity and Specificity, while Area Under Receiver Operating Characteristics (AUROC) \cite{hand2001} measure is found by plotting Recall against False Positive Rate (FPR). Similarly, the Precision and Recall can be combined to form the Area Under Recall Precision Curve (AURPC) \cite{davis2006} index. 

In case of the multi-class classification, a direct extension of the GMean index is available \cite{branco2016}. The multi-class analog of Recall is the Average Class-Specific Accuracy (ACSA) \cite{huang2016learning}. AUROC can be extended for multi-class classification problems by either the One Versus One (OVO) strategy to calculate AUROC-OVO \cite{hand2001}, or by the One Versus All (OVA) strategy to find AUROC-OVA \cite{japkowicz2013asses}. Similarly, the multi-class version of AURPC is called AURPC-OVA \cite{japkowicz2013asses}, as the extension warrants use of the OVA strategy. In the following Table \ref{tab:indexList} we briefly describe the indices which are analysed in detail in the subsequent sections of this article.

\begin{table}[!ht]
    \centering
    \caption{Brief description of the indices discussed in this article (formally detailed in Definition \ref{twoClassDef} and \ref{multiClassDef}).}
    \label{tab:indexList}
    \vspace{0.1cm}
    \scriptsize
    \begin{tabular}{lm{13cm}} \toprule
       Index & Brief description \\ \midrule
       GMean \cite{kubat1997} & Geometric mean of all the class-specific accuracies. Applicable to two-class as well as multi-class classification problems. \\ \midrule
       AUROC \cite{hand2001} & Can be reduced to the arithmetic mean of the class-specific accuracies in a two-class classification problem.\\ \midrule
       Precision \cite{buckland1994} & In a two-class classification problem it is defined as the fraction of true positives to the total number of instances which are classified as positives. \\ \midrule
       AURPC \cite{davis2006} & Reduces to the arithmetic mean of accuracy over the positive class and Precision in a two-class classification problem. \\ \midrule
       ACSA \cite{huang2016learning} & Arithmetic mean of the class-specific accuracies in a multi-class classification problem. \\ \midrule
       AUROC-OVA \cite{japkowicz2013asses} & Direct extension of AUROC for multi-class classification using OVA strategy. \\ \midrule
       AUROC-OVO \cite{hand2001} & Direct extension of AUROC for multi-class classification using OVO strategy. \\ \midrule
       AURPC-OVA \cite{japkowicz2013asses} & Direct extension of AURPC for multi-class classification using OVA strategy.\\ \bottomrule
    \end{tabular}
\end{table}

\subsection{Background} \label{background}
The growing number of classification performance measures inspired the research community to investigate their uniqueness, compare their applicability to class imbalanced problems in general, and evaluate their suitability for specific applications. Studies like \cite{daskalaki2006evaluation, ferri2009, ballabio2018} attempted to empirically find the inter-relation between indices, while observing their behaviour under different scenarios. However, these empirical analyses are dependent on the choice of classifiers as well as the datasets, therefore failing to provide general conclusions. In contrast, a theoretical approach is taken in \cite{Joshi2002} and \cite{liu2007framework} to model the change of an index value with respect to varying class imbalance. However, these preliminary approaches, in addition to being complicated did not consider the possible disparity between the training and test sets. A simpler, more structured framework was proposed by Sokolava and Lapalme \cite{sokolova2007, sokolova2009} which was later extended by Brzezinski \emph{et al.} \cite{brzezinski2018}. They formalized a set of transformation conditions on the confusion matrix \cite{kubat1997} to imitate changes in classification performance as well as alterations in the test set. An index is called invariant (or considered unaffected) by a certain transformation if its value does not change despite the transformation. Luque \textit{et al.} \cite{luque2019} took a different direction by building upon the measure of class imbalance proposed in \cite{Nunez2017} and defining a set of indicators to theoretically analyze the bias of various indices in binary classification problems. Recently, the work done by Brzezinski \emph{et al.} \cite{brzezinski2018} was further extended in \cite{Brzezinski2019} for imbalanced streaming data classification \cite{Yamazaki2007, alaiz2008}. The authors attempted to properly interpret the value returned by an index especially under the effect of the dynamically changing class priors between the training and test sets, which is fairly common for streaming data. However, these works mostly focused on the application-specific suitability of an index. This limits them from discussing on a set of necessary conditions, violation of which may deem the index as undesirable for general use, along with offering any remedial modifications to impose invariance. Moreover, they considered the key dataset properties such as the number of classes as constant. This restricts them from addressing the pivotal role that the altering number of classes may play in distorting an index, a situation common to open set classification problems \cite{rudd2018}. 

\begin{figure}[!t]
    \centering
    \subfloat[Type 1 distortion of index. \label{fig:distort1}]{\includegraphics[width=0.3\textwidth]{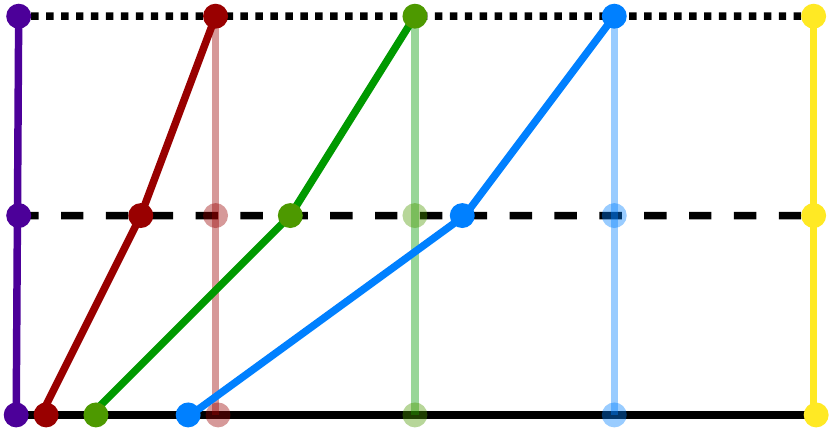}} 
    \hspace{2mm}
    \subfloat[Type 2 distortion of index. \label{fig:distort2}]{\includegraphics[width=0.3\textwidth]{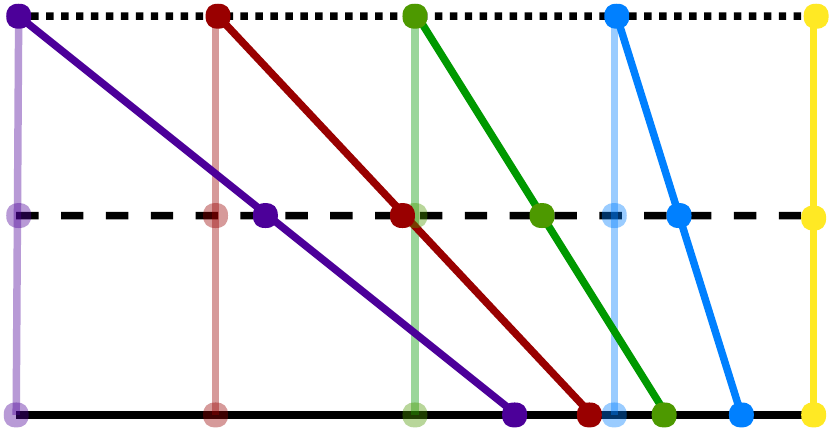}}
    \hspace{2mm}
    \subfloat[Legends. \label{fig:distort-legends}]{\includegraphics[width=0.3\textwidth]{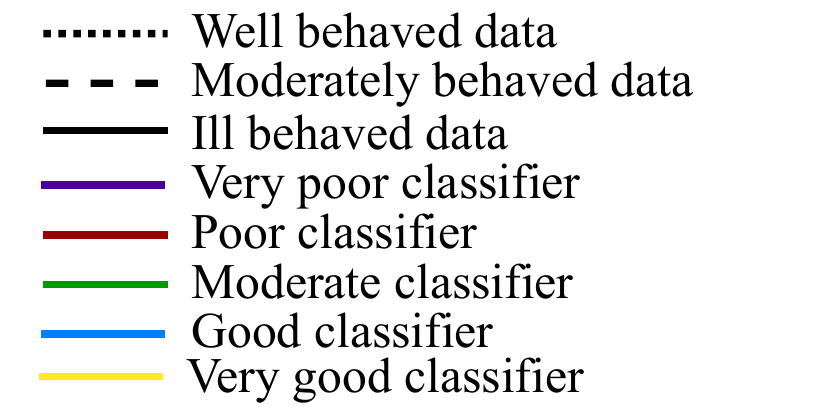}} 
    \caption{Two types of distortions can affect an index while quantifying the performance of classifiers of varying quality over datasets posing diverse degrees of challenge. The complexity of the datasets (plotted by black lines along the horizontal axis) ranges over well behaved (dotted line), moderately behaved (dashed line), and ill behaved (solid line). The quality of the classifier (plotted as colored lines along the vertical axis) varies between very poor (magenta), poor (red), moderate (green), good (blue), and very good (yellow). The ideal behavior of an index is illustrated in the background. (a) Type 1 distortion results in the index becoming increasing warped within its stipulated range. Here the behavior of a dataset is characterized by the variation in the class priors from the training set to those of the test sets. In well behaved data no variation takes place while mild and high amount of disparity is respectively observed for moderately and ill behaved data. (b) Type 2 distortion results in the range of the index becoming progressively smaller. Here number of classes remains constant in a well behaved data, while small and high increase in $C$ respectively indicates a moderately and ill behaved data. Best viewed in color in the electronic version.}
    \label{figMotiv}
\end{figure}

\subsection{Motivation} \label{sec:motiv}
In an attempt to rectify the shortcomings of the existing literature (as discussed in Section \ref{background}), we carry out a systematic theoretical study on the desirable properties of the indices. The presented properties are fundamental in the sense that they ensure the invariance of an index against the following two types of undesirable distortions:

\textbf{Type 1 distortion:} The fraction of representatives from a class in the test set $Q \subseteq X$ (containing $n$ data instances) may not always be similar to that of $P$. However, under such conditions, the value returned by some performance indices (such as Precision \cite{bradley2006}) tends to vary with changes in the number of test points from the different classes. An example of this type of distortion is illustrated in Figure \ref{fig:distort1}, where the mapping within the range of the index (which remains unchanged) becomes increasingly warped. The change in the fraction of representatives between training and test (or validation) sets may happen in a real-world classification problem due to a couple of reasons. Firstly, concept drift \cite{Yamazaki2007, alaiz2008} can result in continuous alterations of class priors (and consequently the degree of imbalance) over time. Such drifting is fairly common in imbalanced streaming data classification problems \cite{Brzezinski2019}, resulting in different extents of class imbalance in training, validation, and test sets. Secondly, prior probability shift between training, validation and test sets also occurs in current large scale benchmark datasets such as LSUN \cite{yu15lsun} and ImageNet \cite{imagenet}, where the the class priors in the training set are not retained in the predefined validation and test sets. It is important to note that remedial measures like stratified cross-validation \cite{LOPEZ2014} cannot be efficiently applied in both of these situations.

\par If an index suffering from the above-mentioned distortion is used during validation, then the classifier will be improperly evaluated and consequently miscalibrated. Further, in streaming data classification, an application may require the classifier to be tested at regular intervals, so that the classifier parameters can be periodically fine-tuned according the latest performance. Here also, use of an index which is susceptible to this first type of distortion may lead to inappropriate judgment about the quality of the classifier and mislead the periodic retraining procedure. 

\textbf{Type 2 distortion:} The range of possible values to be returned by some performance indices (for example, AUROC-OVO as shown in Theorem \ref{theoMultiCon1}) gets diminished with increase in the number of classes in the data. Thus, an index affected by such a distortion may fail to identify the better classifier with decreasing confidence as the number of classes increases, even when the contenders are of diverse quality. In the worst case, on a very large number of classes, due to rounding error a set of classifiers may end up being evaluated as similar, all providing commendable performance instead of reflecting their actual quality. An example of this distortion is also illustrated in Figure \ref{fig:distort2}, where the lower bound of the index gradually increases. 

\begin{figure}[!ht]
    \centering
    \subfloat[\label{fig:type1Best}]{\includegraphics[width=0.3\textwidth]{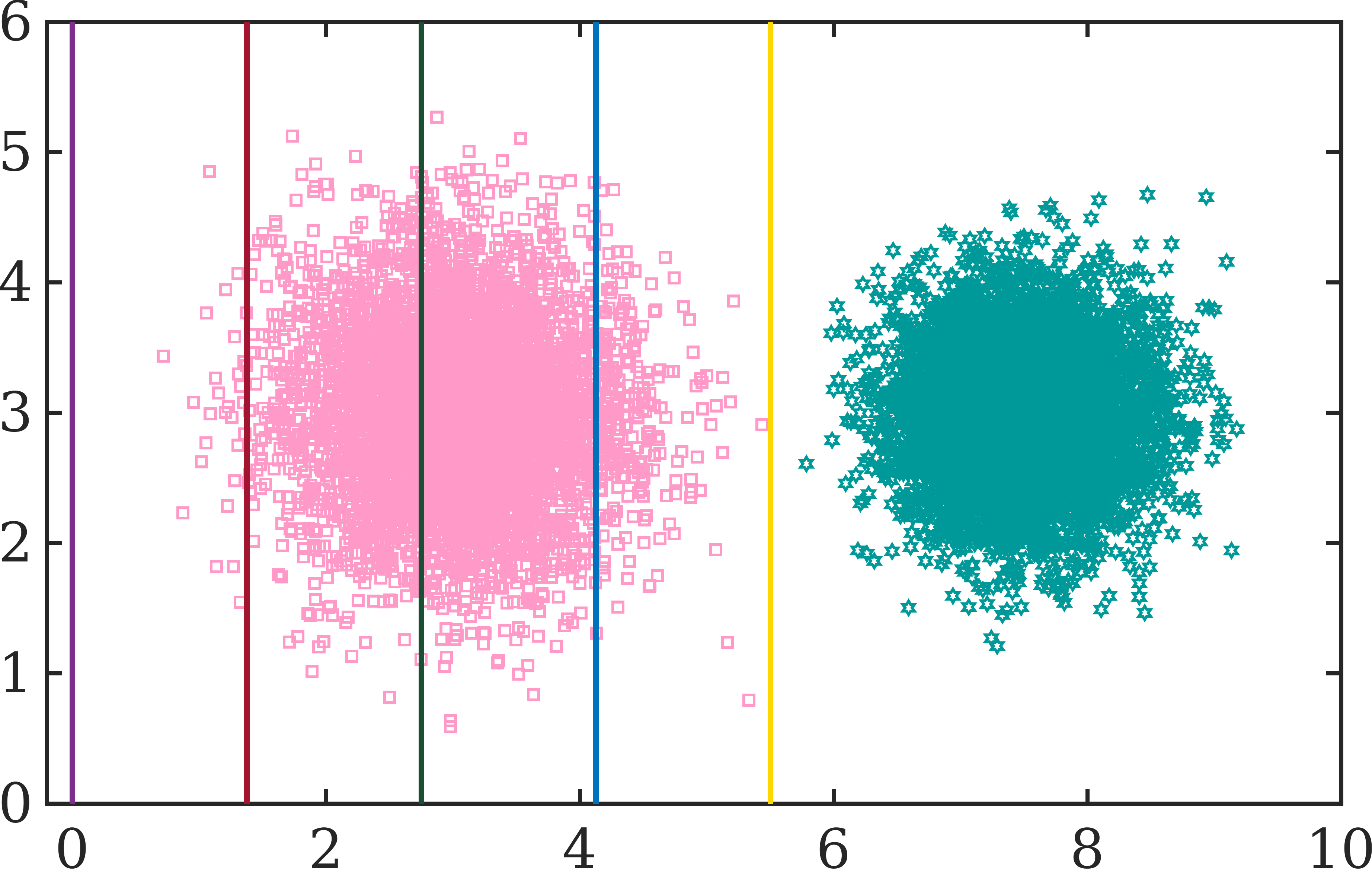}} 
    \hspace{3mm}
    \subfloat[\label{fig:type1Worst}]{\includegraphics[width=0.3\textwidth]{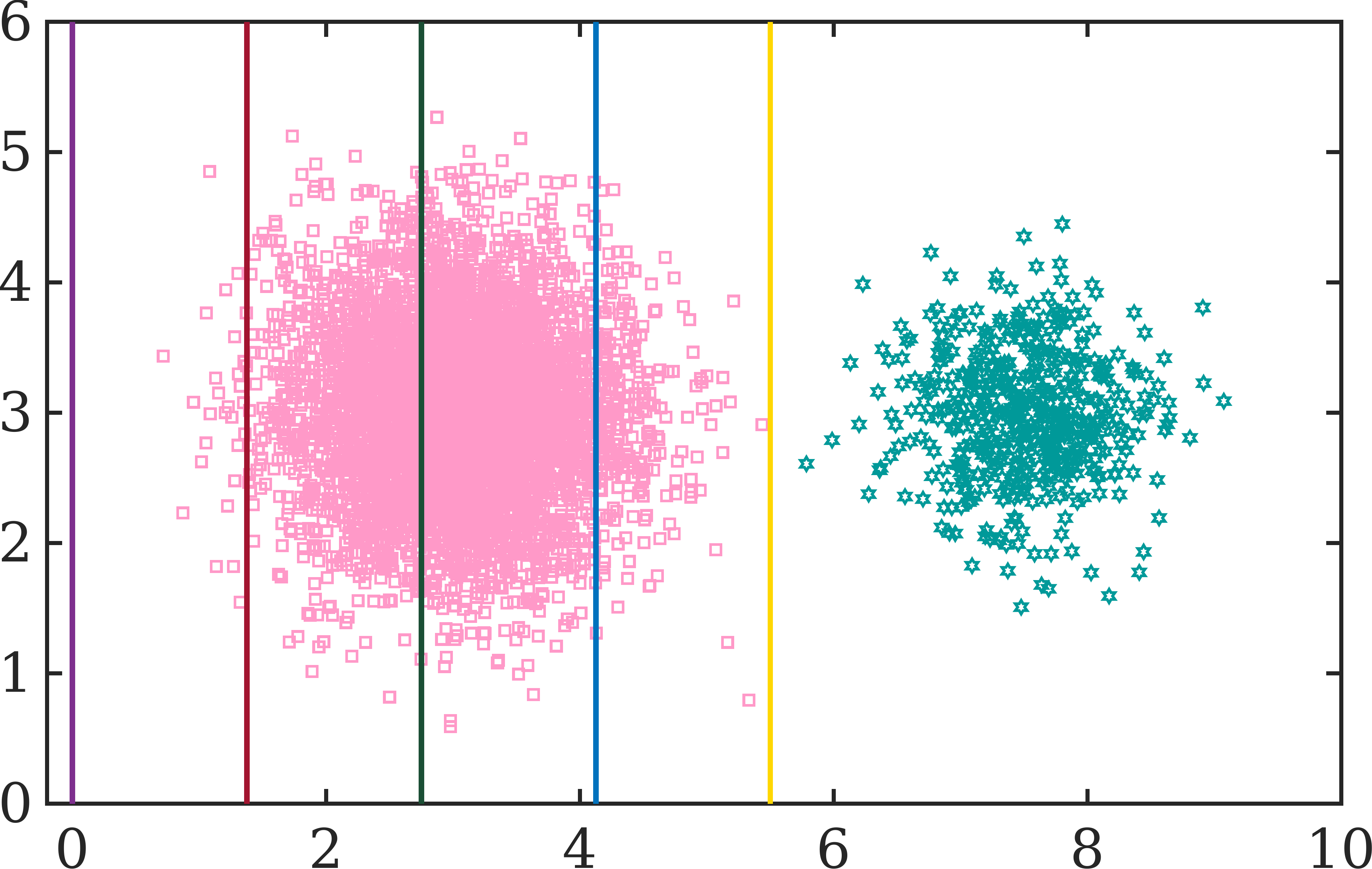}}
    \hspace{3mm}
    \subfloat[\label{fig:type1prec}]{\includegraphics[width=0.32\textwidth]{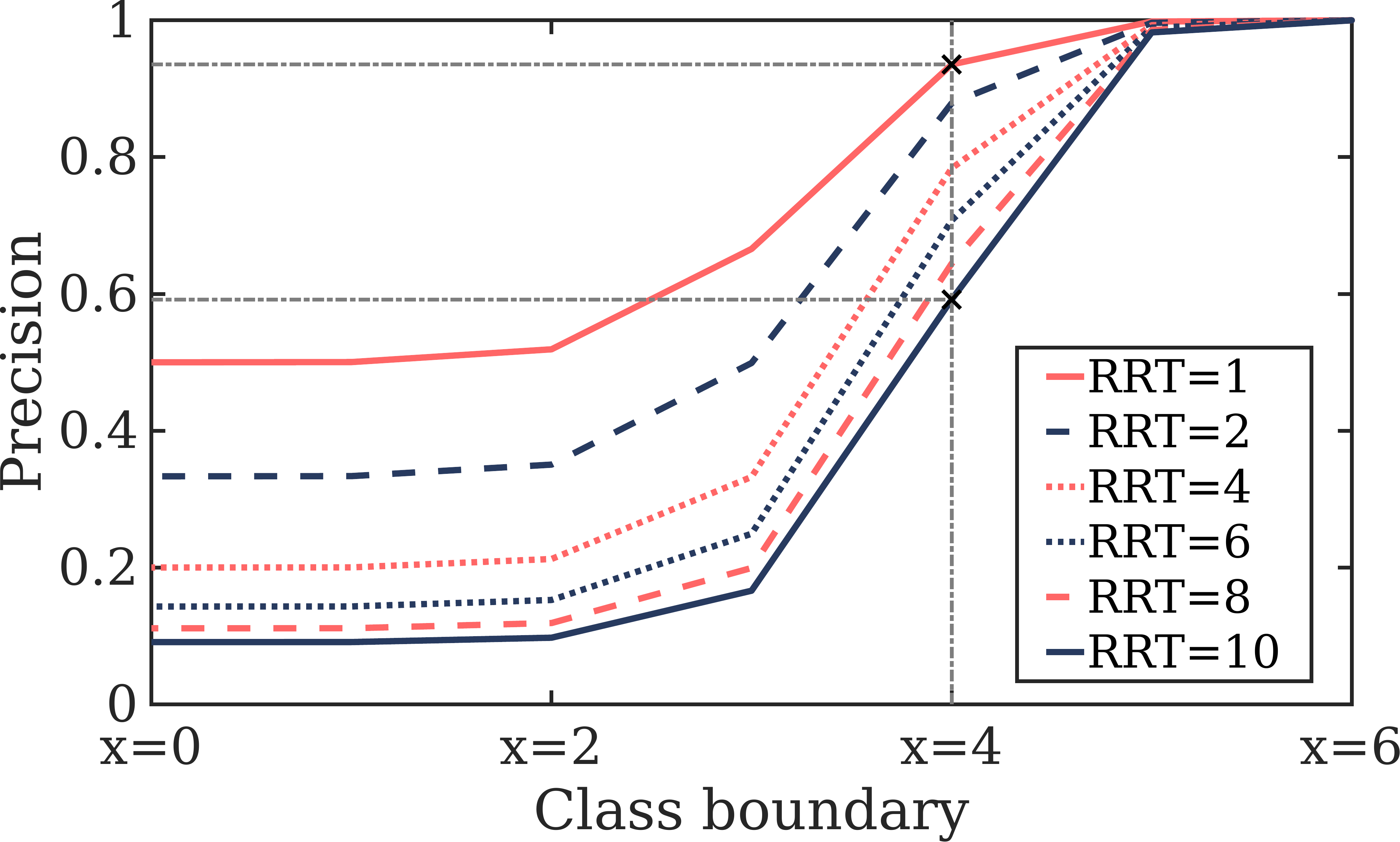}} \\
    \subfloat[\label{fig:type2Best}]{\includegraphics[width=0.3\textwidth]{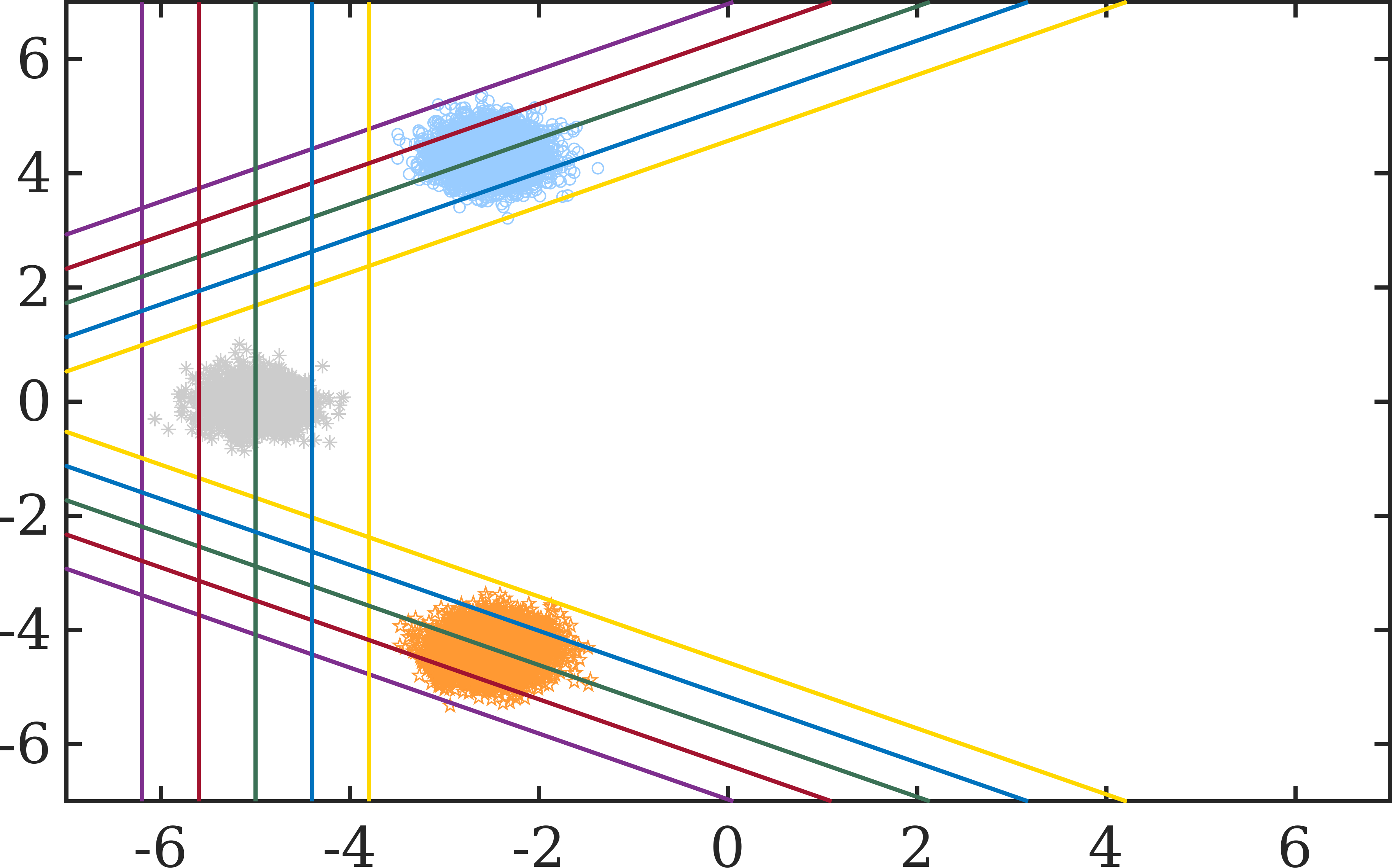}} 
    \hspace{3mm}
    \subfloat[\label{fig:type2Worst}]{\includegraphics[width=0.3\textwidth]{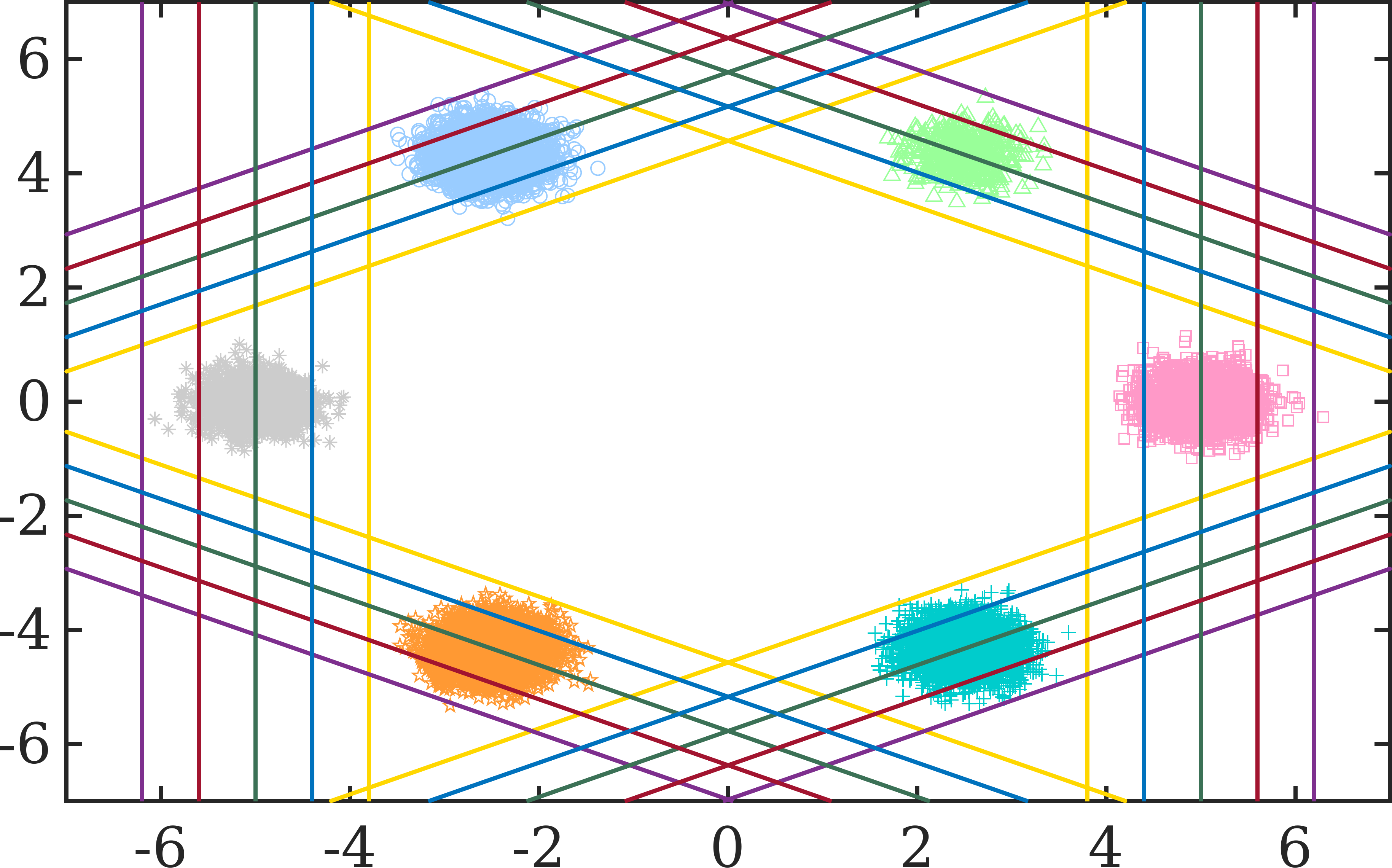}}
    \hspace{3mm}
    \subfloat[\label{fig:type2Auc}]{\includegraphics[width=0.32\textwidth]{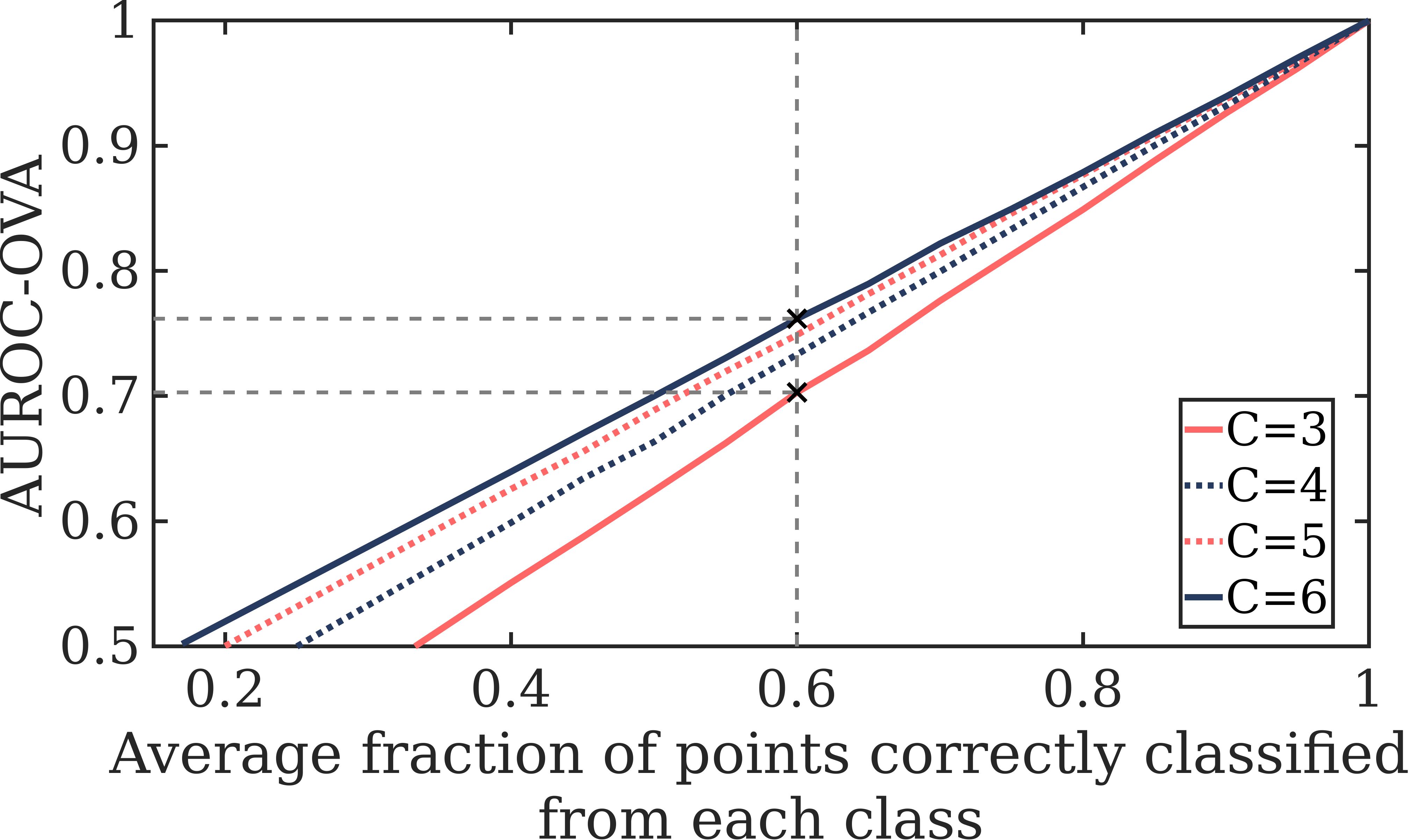}} \\
    \caption{Illustrative example of the two types of distortions; (a)-(c) for Type 1 and (d)-(f) for Type 2. Please note that the quality of the linear classifiers follow the legends as in Figure \ref{figMotiv}. (a) Balanced two class dataset (Well behaved) (b) When RRT is set to 10 by sub-sampling from the class in the right (Ill behaved). (c) Effect of Type 1 distortion on Precision: the value returned by the index deteriorates with increasing RRT, even when the classifier remains the same. (d) Three class dataset (Well behaved) (e) The final six class dataset after adding the rest of three classes on the vertices of the regular hexagon (Ill behaved in the sense of Type 2 distortion). (f) Effect of Type 2 distortion on AUROC-OVA: progressively higher index value is produced for the similar performing classifier while $C$ increases. Best viewed in color in the electronic version.}
    \label{fig:ex}
\end{figure}

\begin{example}
To better illustrate the two types of distortions we present an example in Figure \ref{fig:ex}. For Type 1 distortion, we take a two class dataset\footnote{The construction of all the datasets used in this example is detailed in the supplementary document.} where each class is drawn from a normal distribution. To quantize the level of class imbalance in the test set, we use a measure called Ratio of Representatives in the Test set (RRT), which for a two-class classification problem can be expressed as $\frac{n_{maj}}{n_{min}}$, where $n_{maj}$ and $n_{min}$ are respectively the number of test points from the majority and the minority class (see Definition \ref{def:RRT}). A well behaved test set as shown in Figure \ref{fig:type1Best}, is created by sampling 5000 points from each class. Progressively worse behaved test sets are formed by varying the RRT between 2,4,6,8, and 10. Now as shown in Figures \ref{fig:type1Best} and \ref{fig:type1Worst}, let us shift a linear classifier from the best possible position (which accurately separates the two classes) to the worst (which only perfectly classifies the majority). We measure the Precision of the different classifiers on the varying test sets and plot them in Figure \ref{fig:type1prec}. We can observe that even though the classifier maintains its quality by remaining at a fixed position, the Precision decreases with increasing RRT (for example, at the class boundary $x=4$ the Precision deteriorates from 0.93 to 0.59 when RRT is altered from 1 to 10). The change in Precision is high when the classifier is of poor quality, while the gap progressively closes down with improving performance. 

In case of Type 2 distortion we start with a three class dataset as in Figure \ref{fig:type2Best} where the classes are sampled from normal distributions centered at the three adjacent vertices of a regular hexagon. We gradually increase the number of classes to six in Figure \ref{fig:type2Worst} by similarly sampling on the rest of the three vertices in an anticlockwise order. In each case we start with an OVA ensemble of linear classifiers which performs as worse as an uniformly random assignment and gradually move towards the best which achieves perfect accuracy. As previous we calculate the AUROC-OVA of the different classifiers on the various datasets and plot them in Figure \ref{fig:type2Auc}. We can observe that even when a classifier is accurately classifying the same fraction of points from each class the AUROC-OVA index gradually returns a higher value with the increasing number of classes (for example, when the classifier correctly predicts 60\% points from each class on average, the AUROC-OVA reaches from 0.70 to 0.77 with $C$ altering from 3 to 6). 
\end{example}

Evidently, an evaluation index may suffer from either or both of these distortions with the change in the properties of the dataset (such as number of classes, size of the test set, and extent of class imbalance) even if the classifier retains a consistent performance. Consequently, such types of distortions primarily complicate the interpretation of a value returned by an index. For example, the classifier with moderate performance in Figure \ref{figMotiv} can either be assigned a higher index value or a lower index value (compared to the ideal) depending on the nature of the distortion suffered by the performance indices. Hence, the actual index values cannot be used to properly asses the merit of a classifier. Another issue arises when the difference between the values yielded by a bad classifier and a good one gets diminished to the extent of being ignored due to the rounding of values in practical experiments.

Even when an index is found to be unaffected by the two types of distortions it may still provide a value from which adequate information about the performance of a classifier over all the classes is difficult to extract. This usually occurs in multi-class imbalanced classification problems, where high misclassification in a single class (irrespective of the classifier's performance over the other classes) results in a severely deteriorated index value.

\begin{table}[!ht]
    \centering
    \caption{Summary of contributions made in this article in comparison to existing literature (no references are provided for original contributions).}
    \label{tab:summary}
    \vspace{0.1cm}
    \scriptsize
    \begin{tabular}{m{6cm}C{1.5cm}m{7cm}} \toprule
        Topics & Reference & Our contribution \\ \toprule
        Condition \ref{cond1} & \cite{sokolova2009} & Established as a necessary safeguard against Type 1 distortion. \\ \midrule
        Invariance of Recall, Precision, and AUROC to Condition \ref{cond1} & \cite{sokolova2009} & Re-validated using a mathematical framework. \\ \midrule
        Invariance of GMean, ACSA, AUROC-OVA, AUROC-OVO, and AURPC-OVA to Condition \ref{cond1} & - & Validated using a mathematical framework. \\ \midrule
        Remedial modification of Precision, AURPC, and AURPC-OVA to satisfy Condition \ref{cond1} & \cite{bradley2006} & Validated using a mathematical framework. Established as an effective replacement to the original ones, which satisfy Condition \ref{cond1}. \\ \midrule 
        Remedial normalization of AUROC-OVA to satisfy Condition \ref{cond1} & - & Proposed and validated using a mathematical framework. \\ \midrule
        Condition \ref{cond2} & - & Proposed as a necessary safeguard against Type 2 distortion. \\ \midrule
        Invariance of Recall, Precision, AUROC, GMean, ACSA, AUROC-OVO, AUROC-OVA, and AURPC-OVA to Condition \ref{cond2} & - & Validated using a mathematical framework. \\ \midrule
        Remedial normalization of AUROC-OVO, and AUROC-OVA to satisfy Condition \ref{cond2} & - & Proposed by us and validated using a mathematical framework. \\ \midrule
        Condition \ref{cond3} & - & Proposed to validate the quality of the information returned by an index which satisfies Condition \ref{cond1} and \ref{cond2}. \\ \midrule
        Invariance of GMean, ACSA, and modified AURPC-OVA to Condition \ref{cond3} & - & Validated using a mathematical framework. \\ \bottomrule
    \end{tabular}
\end{table}

\subsection{Contribution}
In this study, we identify two necessary conditions (described in Section \ref{property}) that an index must satisfy to be considered ideal for evaluating the performance of classifiers on imbalanced datasets. Contrary to the prior work, we do not focus on application specific suitability of any index, and aim to propose a set of constraints which will evaluate an index from a more generalized perspective. To elaborate we look at the nature of changes in the data itself which can affect an index. Evidently, it is expected that an index should remain invariant to any changes in the training/test set if the classifier performs uniformly. Therefore, ensuring such can be considered as a fundamental requirement over all other types of secondary consistency checks. In essence, under the assumption that the classifier sustains its performance over each of the classes, we formulate two transformation conditions on the confusion matrix. Invariance to both of these transformations will ensure immunity of an index against the two types of distortions. In Table \ref{tab:summary}, we put our contributions in proper context with the existing works. We further summarize the contributions as follows: 
\begin{enumerate}
    \item The first condition guards against the Type 1 distortion by ensuring invariance of an index with alterations of the size and sample distributions among the different classes in the test test. This condition was first introduced by Sokolova and Lapalme as properties $I_{6}$ and $I_{8}$ in \cite{sokolova2009} (where the former is a special case of the later). However, they were not motivated to evaluate the effects of distortions over the indices. Therefore, their analysis did not elaborate on the implication of the properties or identify them as fundamental constraints. In this article we bridge this gap by establishing this condition as a necessary measure against the Type 1 distortion. Moreover, in Theorems \ref{theo1} and \ref{theoMultiCon1} under the light of the first condition we analytically discuss the properties of GMean, ACSA, AUROC-OVO, AUROCC-OVA, and AURPC-OVA, none of which were covered in the previous studies.
    \item We propose the second condition which deals with the Type 2 distortion assuring invariance to varying number of classes in the test set, as shown in Theorem \ref{theoMultiCon1}.
    \item We show in Theorem \ref{theo2} that contrary to the regular Precision and AURPC indices, the modifications proposed in \cite{bradley2006} are indeed capable of inducing invariance under the first condition. We further propose the normalized variants of AUROC-OVO (which essentially reduces to ACSA) and AUROC-OVA, which offer immunity against the effects of the two types of distortions.
    \item We also propose the third condition to ensure that in a multi-class classification problem, an index which fulfills the two fundamental desirable properties are also capable to provide sufficient information about the classifier's performance over all the classes, even when a single class suffers extremely high misclassification. We show in Theorem \ref{cond3check} that except GMean, both ACSA and AURPC-OVA offer invariance under the third condition. 
\end{enumerate}
In this paper, we also present an empirical analysis on some selected subsets of ImageNet \cite{imagenet} in Section \ref{res} to experimentally validate our theoretical findings and effectiveness of the prescribed remedies. Finally, in Section \ref{discuss}, we present a discussion on the applicability of different indices in an imbalanced classification tasks, and make recommendations as per situation, and subsequently conclude in Section \ref{conclude}.

\section{Desirable Properties for Performance Indices} \label{property}
Various performance evaluating indices depend on the diverse properties (such as imbalance, number of classes etc.) of the training and/or test set to different extents, resulting in improper/ambiguous evaluation of a classifier. This issue can be resolved by defining a set of necessary but not sufficient conditions to ensure the quality of the evaluation by an index. We start with the definition of measures to quantify the extent of class imbalance in the training and test sets. 
\begin{definition}[Datta and Das \cite{datta2018multiobjective}]For a 2-class classification problem the Imbalance Ratio (IR) is defined as the ratio of the number of points in the majority class to that of the minority class in the training set. Analogously, for a $C$-class classification problem IR is calculated as the maximum among all the pairwise IRs (represented by the set $\mathbb{I}=\{\frac{p_{i}}{p_{j}}|i, j \in \mathcal{C}; i \neq j \}$, where $p_{i}$ is the number of training points from the $i^{th}$ class) among the $C$ classes (i.e. IR = $\max{\mathbb{I}}$). Therefore, $P$ can be considered as imbalanced if $\text{IR} > 1$.
\end{definition}
However, a classifier is trained on a single training set $P$ with a fixed predefined $\mathbb{I}$, whereas all possible test sets might not follow a distribution of the representatives among the classes (i.e. $\mathbb{I}$) similar to $P$. Therefore, analogous to IR we define RRT for quantifying the ratio of representatives among the classes in the test set. 
\begin{definition}
\label{def:RRT}
For a 2-class classification problem, the RRT is defined as the proportion of the number of points from the majority class to that of the minority class, where the majority and the minority classes are named according to the training set. This definition can be extended for the $C$-class classification case in a manner similar to IR, where the set of a pairwise ratio of the number of the data instances among the $C$ classes is denoted by $T$, i.e. RRT = $\max{T}$.
\end{definition}
Performance of the classifier on a $C$-class classification problem can be expressed in the form of a matrix called the \emph{confusion matrix}, which is defined as follows:
\begin{definition}[Kubat \textit{et al.} \cite{kubat1997}]
\label{confmat}
A confusion matrix over a test set $Q$ for a $C$-class classification problem can be defined as $M_{C}=[m_{ij}]_{C \times C}$, where $m_{ij}$ represents the number of points which actually belongs to $i^{th}$ class but are predicted as a member of class $j$, for all $i, j \in \mathcal{C}$. Thus, the diagonal elements i.e. $m_{ii}$ are those instances of class $i$ which are correctly classified while the rest are different misclassifications. Evidently, each entry in the confusion matrix must be a non-negative integer i.e. $m_{ij} \in \mathbb{Z}^{+} \cup \{0\}; \; \forall i, j \in \mathcal{C}$
\end{definition}
There are some important properties of the confusion matrix which we detail in the following discussion.

\textit{Property 1:} The sum of entries in the $i^{th}$ row of the confusion matrix is denoted by $n_i$ (i.e. $\sum_{j=1}^{C}m_{ij}=n_{i}; \; \forall i \in \mathcal{C}$), which is the number of test points belonging to the $i^{th}$ class. We assume that $n_{i}>0$, as there should be at least one point from each class in the test set. 

\textit{Property 2:} The sum of entries in the $i^{th}$ column of the confusion matrix is denoted by $k_i$ (i.e. $\sum_{j=1}^{C}m_{ji}=k_{i}; \; \forall i \in \mathcal{C}$), which is the number of test points predicted as $i^{th}$ class.

\textit{Property 3:} The total number of test points $n=\sum_{i=1}^{C}\sum_{j=1}^{C}m_{ij}; \; \forall i,j \in \mathcal{C}$. 

\textit{Property 4:} In case of two-class classification, the entries of $M_{2}$ are specially named, as True Positive (TP), False Positive (FP), True Negative (TN), and False Negative (FN), when the test instances from the majority and the minority classes are respectively labelled as -1 (negative) and +1(positive). Therefore, $M_{2}$ can be formally represented as:
\begin{equation}
\label{conf2}
    M_{2}=
    \begin{bmatrix}
        TP & FN \\
        FP & TN \\
    \end{bmatrix}.
\end{equation}

Before proceeding further we need to describe our primary assumption based on which the following theory will be built. 
\begin{assumption} \label{asm1}
The class-specific performance (the fraction of correct classification as well as the proportion of misclassification to each of the other classes) of a classifier remains the same over any random subset of the dataset.
\end{assumption}

The class-specific performance of a classifier can be considered as an equivalence relation, which can partition the set $\mathbb{M}_{C}$ containing all possible $C$-class confusion matrices into some equivalence classes. In any of these equivalence classes, the class-specific performance of a classifier remains constant over all the classes. To elaborate, given two confusion matrices say $M_{C}$ and $M'_{C}$\footnote{Evidently, $M'_{C}=[m'_{ij}]_{C\times C}$, $\sum_{i=1}^{C}\sum_{j=1}^{C}m'_{ij}=n'$, $\sum_{j=1}^{C}m'_{ij}=n'_{i}$, and $\sum_{j=1}^{C}m'_{ji}=k'_{i}$ for all $i \in \mathcal{C}$.}, if the equivalence relation ${m_{ij}}/{n_{i}}={m'_{ij}}/{n'_{i}}$, satisfies for all $i,j \in \mathcal{C}$, then they can be considered as members of the same equivalence class, i.e. $M_{C} \sim M'_{C}$. In other words the equivalence property essentially corresponds to constant performance by a classifier or formally represents Assumption \ref{asm1}. Using the notion of the confusion matrix, we can now formally define a performance evaluation index as a function $f$ mapping from the set of all possible confusion matrices $\mathbb{M}_{C}$ to a real scalar quantity. Such a representation is important as it helps us define some functionals to formally describe our proposed conditions, in the following manner.
\begin{condition}
\label{cond1}
The value of an index should not be dependent on RRT, if the classifier performs equivalently, i.e. 
\begin{align*}
&\mathcal{V}_{M_{C}}(f)=\mathcal{V}_{M'_{C}}(f); \; \forall M_{C} \sim M'_{C},
\end{align*}
where $\mathcal{V}_M$ is a functional evaluating the index $f$ on the confusion matrix $M$.
\end{condition}

As an extension of the work by Sokolova and Lapalme \cite{sokolova2009}, in this article we propose Condition \ref{cond1} as a necessary measure against the Type 1 distortion, violation of which may alter the value of an index with the changes of RRT in the test set even when the classifier remains the same.

\begin{condition}
\label{cond2}
The lower and the upper bounds of an index $f$ should not be dependent on the number of classes, i.e. \begin{align*}
    &\mathcal{L}_{\mathbb{M}_{C}}(f)=\mathcal{L}_{\mathbb{M}_{C+1}}(f) \text{ } \forall C \in \mathbb{Z}^{+}\setminus \{1\},\\
    \text{and }&\mathcal{U}_{\mathbb{M}_{C}}(f)=\mathcal{U}_{\mathbb{M}_{C+1}}(f) \text{ } \forall C \in \mathbb{Z}^{+}\setminus \{1\},
\end{align*}
where, $\mathbb{M}_{C}$ and $\mathbb{M}_{C+1}$ are respectively the sets of all possible $C$-class and $(C+1)$-class confusion matrices. Moreover, $\mathcal{L}$ and $\mathcal{U}$, are two functionals of $f$, respectively calculating the minimum and maximum value of $f$ over all $\mathbb{M}_{C}$ and $\mathbb{M}_{C+1}$.
\end{condition}
Condition \ref{cond2} ensures that under the assumption of a consistent performance by a classifier, the value of an index should not be biased to differing number of classes in the test set. 

If we consider a $C$-class confusion matrix, where $m_{ii}/n_{i}=\epsilon$, only for the $i^{th}$ class ($i \in \mathcal{C}$) while $m_{jj}/n_{j}\geq (1-\epsilon)$ for all the other classes ($j \in \mathcal{C}\setminus \{i\}$), and $\epsilon=\frac{1}{C}$, then all such matrices form a set $\mathcal{M}_{C}(i) \subset \mathbb{M}_{C}$. In other words, $\mathcal{M}_{C}(i)$, is the set of all such $C$-class confusion matrices where the classifier performed extremely poor only on the $i^{th}$ class. 
\begin{condition}
\label{cond3}
An index $f$ while evaluating a multi-class classifier should not be biased towards the misclassification of a single class, i.e. $\mathcal{W}_{\mathcal{M}_{C}(i)}(f) > \mathcal{L}_{\mathbb{M}_{C}}(f)$, where $\mathcal{W}$ is a functional which calculates the limit of $f$, as $\epsilon \rightarrow 0^{+}$.
\end{condition}

In other words, Condition \ref{cond3} ensures that the value returned by the index will not excessively degrade if a single class is almost entirely misclassified in a multi-class classification problem. Violating Condition \ref{cond3} will lead the index to lose information about the classifier's performance on all the other classes. Evidently, index failing to satisfy Condition \ref{cond3} will be unable to distinguish between two classifiers, one of which achieves good class-specific accuracies on all but one class, while the other achieves high misclassification on all.   

Before proceeding further, we list down the various notations which will be used throughout the rest of this article in Table \ref{tab:notations}.

\begin{table}[!ht]
    \centering
    \caption{List of notations}
    \vspace{0.1cm}
    \label{tab:notations}
    \scriptsize
    \begin{tabular}{ll} \toprule 
    $X, P, Q$ & Dataset, training set, and test set respectively. \\
    $p_{i}$ $(n_{i})$ & Number of training (test) points in the $i^{th}$ class. \\
    $m_{ij}$ & Number of test points from $i^{th}$ class classified as $j^{th}$. \\
    $f$ & A classification performance evaluation index. \\
    $M_{2}$ $(M_{C})$ & Two-class (multi-class) confusion matrix. \\
    $\mathbb{M}_{C}$ & Set of all multi-class confusion matrices. \\
    $\gamma_{2}$ $(\gamma_{C})$ & Two-class (multi-class) GMean index. \\
    $\rho$, $\rho_{o}$, $\rho_{a}$ & AUROC, AUROC-OVO, and AUROC-OVA index. \\
    $\alpha$ & Average class-specific accuracy index. \\
    $\kappa$, $\kappa_{a}$ & AURPC and AURPC-OVA index. \\
    $\mathcal{L}$ $(\mathcal{U})$ & Functional for the lower (upper) bound of an index. \\ 
    $\mathcal{M}_{C}(i)$ & Subset of $\mathbb{M}_{C}$, where $i^{th}$ class suffers high misclassification. \\
    $\mathcal{W}$ & Functional for calculating the limit of an index $f$. \\ \bottomrule
    \end{tabular}
\end{table}

\section{Analysis of the two-class performance evaluation indices}
In this section, we analyze the characteristics of four indices, namely GMean, AUROC \footnote{For mathematical simplicity we have restricted ourselves to the discrete version of the index, which is popularly used in practice.}, Precision, and AURPC which are used to evaluate the performance of a classifier in presence of class imbalance for a two-class classification problem. We only require to validate if the indices satisfy Condition \ref{cond1} as the other is only applicable for multi-class classification. 

\begin{definition}
\label{twoClassDef}
For a two-class classification problem, given a confusion matrix $M_{2}$ as in (\ref{conf2}),
\begin{enumerate}
    \item The GMean index, denoted by $\gamma_{2}$ is defined as:
    \begin{equation}
    \label{gm2}
        \gamma_{2}(M_{2})={\left(\left(\frac{TP}{TP+FN}\right)\left(\frac{TN}{FP+TN}\right)\right)}^{\frac{1}{2}}.
    \end{equation}
    \item The AUROC index, denoted by $\rho$ is defined as:
    \begin{equation}
    \label{auc2}
        \rho(M_{2})=\frac{1}{2}\left(\frac{TP}{TP+FN}+\frac{TN}{FP+TN}\right).
    \end{equation}
    \item The Precision index, denoted by $\zeta$ is defined as:
    \begin{equation}
    \label{prec}
        \zeta(M_{2})=\frac{TP}{TP+FP}.
    \end{equation}
    \item The AURPC index, denoted by $\kappa$ is defined as:
    \begin{equation}
    \label{rp2}
        \kappa(M_{2})=\frac{\beta(M_{2})+\zeta(M_{2})}{2},
    \end{equation}
    where, $\beta(M_{2})=TP/(TP+FN)$ is the Recall.  
\end{enumerate}
\end{definition}
Evidently, the formal definitions of the indices do correspond to their pedagogical description in Table \ref{tab:indexList}. We may now proceed to analyzing the behavior of the indices under the light of Condition \ref{cond1} in the following theorem. 

\begin{theorem}
\label{theo1}
For a two-class classification problem, given two confusion matrices $M_{2}$ and $M'_{2}$, the following statements can shown to be true if $M_{2} \sim M'_{2}$,
\begin{enumerate}
    \item The index $\gamma_{2}$ satisfies Condition \ref{cond1}. 
    \item The index $\rho$ satisfies Condition \ref{cond1}. 
    \item The index $\zeta$ does not satisfy Condition \ref{cond1}.
    \item The index $\kappa$ does not satisfy Condition \ref{cond1}.
\end{enumerate}
\end{theorem}
\begin{proof}
Let us define $M_{2}$, as in (\ref{conf2}), while $M'_{2}$ can be constructed as,
\begin{equation*}
    M'_{2}=
    \begin{bmatrix}
        b_{1}TP & b_{1}FN \\
        b_{2}FP & b_{2}TN \\
    \end{bmatrix},
\end{equation*}
where $b_{1}, b_{2} \in \mathbb{R^{+}}$, $b_{1} \neq b_{2}$ and $m'_{ij} \in \mathbb{Z^{+}}, \forall i, j \in \{1, 2\}$. Such a form of $M'_{2}$, will ensure that ${m_{ij}}/{n_{i}}={m'_{ij}}/{n'_{i}}$, is satisfied for all $i,j \in \{1, 2\}$, or $M_{2} \sim M'_{2}$. With this initial setup we start the proof of the first statement by finding the value of $\mathcal{V}_{M'_{2}}(\gamma_{2})$, following (\ref{gm2}):
\begin{align*}
    \mathcal{V}_{M'_{2}}(\gamma_{2})&={\left(\left(\frac{b_{1}TP}{b_{1}TP+b_{1}FN}\right)\left(\frac{b_{2}TN}{b_{2}FP+b_{2}TN}\right)\right)}^{\frac{1}{2}}, \\ \nonumber
    &={\left(\left(\frac{TP}{TP+FN}\right)\left(\frac{TN}{FP+TN}\right)\right)}^{\frac{1}{2}}=\mathcal{V}_{M_{2}}(\gamma_{2}).
\end{align*}
Therefore, the $\gamma_{2}$ index satisfies Condition \ref{cond1}.

The second statement can be proved in a similar manner by starting from $\mathcal{V}_{M'_{2}}(\rho)$ using (\ref{auc2}),
\begin{align*}
    \mathcal{V}_{M'_{2}}(\rho)&=\frac{1}{2}\left(\frac{b_{1}TP}{b_{1}(TP+FN)}+\frac{b_{2}TN}{b_{2}(FP+TN)}\right),\\ \nonumber
    &=\frac{1}{2}\left(\frac{TP}{TP+FN}+\frac{TN}{FP+TN}\right)=\mathcal{V}_{M_{2}}(\rho).
\end{align*}
This proves the second statement.

Similarly, we show the third statement to be true by calculating $\mathcal{V}_{M'_{2}}(\zeta)$ as per (\ref{prec}),
\begin{equation}
\label{zetaNew}
    \mathcal{V}_{M'_{2}}(\zeta)=\frac{b_{1}TP}{b_{1}TP+b_{2}FP}.
\end{equation}
From (\ref{zetaNew}) it is evident that $\mathcal{V}_{M'_{2}}(\zeta)$ can be equal to $\mathcal{V}_{M_{2}}(\zeta)$, only when $b_{1}=b_{2}$, which implies that $\zeta$ violates Condition \ref{cond1}.

Finally, we prove the fourth statement by finding the value of $\mathcal{V}_{M'_{2}}(\kappa)$ according to (\ref{rp2}),
\begin{align}
\label{aurpc2}
    \mathcal{V}_{M'_{2}}(\kappa)&=\frac{b_{1}TP}{b_{1}(TP+FN)}+\frac{b_{1}TP}{b_{1}TP+b_{2}FP}, \nonumber \\
    & =\frac{TP}{TP+FN}+\frac{b_{1}(TP)}{b_{1}(TP)+b_{2}(FP)}.
\end{align}
Therefore, from (\ref{aurpc2}), we can conclude in a manner similar to (\ref{zetaNew}) that $\mathcal{V}_{M'_{2}}=\mathcal{V}_{M_{2}}$ only holds when $b_{1}=b_{2}$, thus completing the proof.
\end{proof}

From Theorem \ref{theo1}, we can see that while GMean and AUROC indices satisfy Condition \ref{cond1}, the Precision and AURPC indices do not, thus being susceptible to RRT. In other words, both of Precision and AURPC may evaluate a good classifier as a poor choice, with the increase in RRT, even when the class-specific performances are retained. This is due to the increasing number of test points from the majority class which considerably increases FP. This was first observed by Bradley \emph{et al.} \cite{bradley2006}, who proposed a solution by incorporating the class priors in the definition of Precision. In their modified definition of Precision (and consequently AURPC) the direct use of FP is replaced with the ratio of false positives to the number of majority instances. The modified Precision and AURPC, respectively called mPrecision and mAURPC, are described in the following definition.
\begin{definition}[Bradley \emph{et al.} \cite{bradley2006}]
For a two-class classification problem, given a confusion matrix $M_{2}$ as in (\ref{conf2}), where $TP+FN=n_{1}$ and $FP+TN=n_{2}$,
\begin{enumerate}
    \item The mPrecision index, denoted by $\hat{\zeta}$ is defined as:
    \begin{equation}
    \label{mPrec}
        \hat{\zeta}(M_{2})=\frac{TP/n_{1}}{(TP/n_{1})+(FP/n_{2})}.
    \end{equation}
    \item The mAURPC index, denoted by $\hat{\kappa}$ is defined as:
    \begin{equation}
    \label{mRp2}
        \hat{\kappa}(M_{2})=\frac{1}{2}(\beta(M_{2})+\hat{\zeta}(M_{2})).
    \end{equation}
\end{enumerate}
\end{definition}

\begin{theorem}
\label{theo2}
For a two-class classification problem, given two confusion matrices $M_{2}$, and $M'_{2}$, the following statements can shown to be true, if $M_{2} \sim M'_{2}$,
\begin{enumerate}
    \item The index $\hat{\zeta}$ satisfies Condition \ref{cond1}.
    \item The index $\hat{\kappa}$ satisfies Condition \ref{cond1}.
\end{enumerate}
\end{theorem}
\begin{proof}
Given $M_{2}$, as in (\ref{conf2}), we construct $M'_{2}$, such that $M_{2} \sim M'_{2}$, in a manner similar to Theorem \ref{theo1}. Then to prove the first statement we proceed by calculating $\mathcal{V}_{M'_{2}}(\hat{\zeta})$ using (\ref{mPrec}).
\begin{align*}
    \mathcal{V}_{M'_{2}}(\hat{\zeta})&=\frac{b_{1}TP/b_{1}n_{1}}{b_{1}TP/b_{1}n_{1}+b_{2}FP/b_{2}n_{2}}, \nonumber \\
    &=\frac{TP/n_{1}}{(TP/n_{1})+(FP/n_{2}}=\mathcal{V}_{M_{2}}(\hat{\zeta}).
\end{align*}
This completes the proof of first statement.

We prove the second statement by finding $\mathcal{V}_{M'_{2}}(\hat{\kappa})$, which from (\ref{mPrec}), and (\ref{mRp2}) can also be written as
\begin{align*}
    \mathcal{V}_{M'_{2}}(\hat{\kappa})&=\frac{b_{1}TP}{2b_{1}n_{1}}+    \frac{b_{1}TP/b_{1}n_{1}}{2b_{1}TP/b_{1}n_{1}+2b_{2}FP/b_{2}n_{2}} \nonumber \\
    &=\frac{TP}{2n_{1}}+\frac{TP/n_{1}}{(TP/2n_{1})+(FP/2n_{2})}=\mathcal{V}_{M_{2}}(\hat{\kappa}).
\end{align*}
Thus, the second statement is proved, completing the proof of this Theorem.
\end{proof}

From Theorem \ref{theo2} we can conclude that the proposed modification of Precision and AURPC can improve their immunity over RRT by satisfying Condition \ref{cond1}, and in the process will be able to better evaluate a classifier. 

\section{Analysis of the multi-class performance evaluation indices}
In this section, we will define five multi-class evaluation indices, namely GMean, ACSA, AUROC-OVO, AUROC-OVA, and AURPC-OVA. Similar to the two-class indices we present an analysis in the perspective of the first two conditions and prescribe modification/normalization as per requirement. 
\begin{definition}
\label{multiClassDef}
Given a $C$-class confusion matrix $M_{C}$ as defined in Definition \ref{confmat}, 
\begin{enumerate}
    \item The GMean index, denoted by $\gamma_{C}$ is defined as:
    \begin{equation}
    \label{gmm}
        \gamma_{C}(M_{C})=\left(\prod_{i=1}^{C}{\frac{m_{ii}}{n_{i}}}\right)^{\frac{1}{C}}. 
    \end{equation}
    \item The ACSA index, denoted by $\alpha$ is defined as:
    \begin{equation}
    \label{acsa}
        \alpha(M_{C})=\frac{1}{C}\sum_{i=1}^{C}\frac{m_{ii}}{n_{i}}.
    \end{equation}
    \item The AUROC-OVO index, denoted by $\rho_{o}$ is defined as:
    \begin{equation}
    \label{aucovo}
        \rho_{o}(M_{C})=\frac{1}{2C}\sum_{i=1}^{C}\Bigg(1+\frac{m_{ii}}{n_{i}}-\sum_{\substack{j=1 \\ j \neq i}}^{C} \frac{m_{ji}}{(C-1)n_{j}}\Bigg).
    \end{equation}
    \item The AUROC-OVA index, denoted by $\rho_{a}$ is defined as:
    \begin{equation}
    \label{aucova}
        \rho_{a}(M_{C})=\frac{1}{2C}\sum_{i=1}^{C}\left(1+\frac{m_{ii}}{n_{i}}-\frac{k_{i}-m_{ii}}{n-n_{i}}\right).
    \end{equation}
    \item The AURPC-OVA index, denoted by $\kappa_{a}$ is defined as:
    \begin{equation}
    \label{rpm}
        \kappa_{a}(M_{C})=\frac{1}{2C}\sum_{i=1}^{C}\left(\frac{m_{ii}}{k_{i}}+\frac{m_{ii}}{n_{i}}\right).
    \end{equation}
\end{enumerate}
\end{definition}

Similar to the two-class case, here also the indices reflect their informal description from Table \ref{tab:indexList} to their mathematical definition. Before proceeding further, we need to first prove three supporting lemmas, which respectively comment on the range of ACSA index, and highlights the key properties of AUROC-OVO and AUROC-OVA. 
\begin{lemma}
\label{supp:lem1}
In a $C$-class classification problem the value of index $\alpha$ lies between 0, and 1.
\end{lemma}
\begin{proof}
According to Definition \ref{confmat}, in a $C$-class confusion matrix $0 \leq m_{ii}/n_{i} \leq 1$, for all $i \in \mathcal{C}$. Using this and the definition of $\alpha$ in (\ref{acsa}), we can conclude that $0 \leq \alpha \leq 1$. Specifically, $\alpha=0$, when the classifier wrongly classified every test point i.e. $m_{ii}=0$, for all $i \in \mathcal{C}$, and $\alpha=1$, if the classifier correctly predicts the class label for each member of the test set. 
\end{proof}

\begin{lemma}
\label{lem1}
The value $\rho_{o}(M_{C})$ can be expressed as a linear function of $\alpha(M_{C})$, with a constant coefficient and bias both of which are dependent on $C$, as follows: 
\begin{equation}
\label{aucovoCond21}
    \rho_{o}(M_{C})=\frac{C}{2(C-1)}\alpha(M_{C})+\frac{C-2}{2(C-1)}.
\end{equation}
Moreover, the lower bound of $\rho_{o}(M_{C})$ can be expressed as $\mathcal{L}_{\mathbb{M}_{C}}(\rho_{o})=\frac{C-2}{2(C-1)}$, while the upper bound $\mathcal{U}_{\mathbb{M}_{C}}(\rho_{o})=1$.
\end{lemma}
\begin{proof}
We first start with a $M_{C} \in \mathbb{M}_{C}$, then following from (\ref{aucovo}) after some algebraic manipulation we express $\rho_{o}(M_{C})$ as:
\begin{align}
\label{supp:aucovoCond21}
    \rho_{o}(M_{C})&=\frac{1}{2}+\frac{1}{2C}\sum_{i=1}^{C}\frac{m_{ii}}{n_{i}}-\frac{1}{2C(C-1)}\sum_{i=1}^{C}\sum_{j \in \mathcal{C} \setminus \{i\}} \frac{m_{ji}}{n_{j}}, \nonumber \\
    &=\frac{1}{2}+\frac{1}{2C}\sum_{i=1}^{C}\frac{m_{ii}}{n_{i}}-\frac{1}{2C(C-1)}\sum_{i=1}^{C}\frac{n_{i}-m_{ii}}{n_{i}}, \nonumber \\
    &=\frac{1}{2}+\frac{1}{2C}\sum_{i=1}^{C}\frac{m_{ii}}{n_{i}}-\frac{1}{2(C-1)}\Big(1-\frac{1}{C}\sum_{i=1}^{C}\frac{m_{ii}}{n_{i}}\Big), \nonumber \\
    &=\frac{1}{2}+\frac{\alpha(M_{C})}{2}-\frac{1}{2(C-1)}(1-\alpha(M_{C})), \nonumber \\
    &=\frac{C}{2(C-1)}\alpha(M_{C}) +\frac{C-2}{2(C-1)}.
\end{align}
Interestingly, from (\ref{supp:aucovoCond21}) we can conclude that for a given $C$, the index $\rho_{o}(M_{C})$ can be expressed as a linear function of $\alpha(M_{C})$, with a constant coefficient and a bias. Now $M_{C}$ is an arbitrary matrix belonging to the set $\mathbb{M}_{C}$. Therefore, we can safely extend (\ref{supp:aucovoCond21}) to:
\begin{equation}
\label{aucovoCond22}
    \mathcal{L}_{\mathbb{M}_{C}}(\rho_{o})=\frac{C}{2(C-1)}\mathcal{L}_{\mathbb{M}_{C}}(\alpha) +\frac{C-2}{2(C-1)},
\end{equation}
\begin{equation}
\label{aucovoCond23}
    \text{and, }\mathcal{U}_{\mathbb{M}_{C}}(\rho_{o})=\frac{C}{2(C-1)}\mathcal{U}_{\mathbb{M}_{C}}(\alpha) +\frac{C-2}{2(C-1)}.
\end{equation} 
Plugging the values of $\mathcal{L}_{\mathbb{M}_{C}}(\alpha)$, $\mathcal{U}_{\mathbb{M}_{C}}(\alpha)$ from Lemma \ref{supp:lem1}, respectively in (\ref{aucovoCond22}), and (\ref{aucovoCond23}) we obtain.
\begin{equation*}
    \mathcal{L}_{\mathbb{M}_{C}}(\rho_{o})=\frac{C-2}{2(C-1)} \;
    \text{and }\; \mathcal{U}_{\mathbb{M}_{C}}(\rho_{o})=1,
\end{equation*} 
which completes the proof.
\end{proof}

\begin{lemma}
\label{lem2}
If we assume for simplicity, without loss of generality that $n_{1} \leq n_{2} \cdots \leq n_{C}$, then the lower bound of $\rho_{a}(M_{C})$ can be expressed as
\begin{equation}
    \mathcal{L}_{\mathbb{M}_{C}}(\rho_{a})=\frac{1}{2C}\left( C-1-\frac{n_{C}}{n-n_{C-1}}\right)
\end{equation}
while the upper bound $\mathcal{U}_{\mathbb{M}_{C}}(\rho_{a})=1$.
\end{lemma}
\begin{proof}
If the classifier misclassifies all of the test points then $m_{ii}=0, \forall i \in \mathcal{C}$. However, as evident from (\ref{aucova}) the value of $\mathcal{L}_{\mathbb{M}_{C}}(\rho_{a})$ is also dependent on the actual predictions as the the cost of misclassification to all the classes are not equal. To elaborate, we take a $C$-class confusion matrix $M_{C}$, and construct $M'_{C}$, such that $M_{C}+\Delta=M'_{C}$, where, $\Delta=[\delta_{ij}]_{C \times C}$, $\sum_{i=1}^{C}\delta_{ji}=0$, $m'_{ij}=m_{ij}+\delta_{ij}\geq 0$, and $\sum_{i=1}^{C}\delta_{ij}=\bar{k}_{i}$ $\forall i, j \in \mathcal{C}$, while conserving $n_{i}, \forall i \in \mathcal{C}$, and $n$. Now, $\rho_{a}(M_{C})-\rho_{a}(M'_{C})$ can be calculated as
\begin{equation}
\label{aucovaCond21}
      \rho_{a}(M'_{C})-\rho_{a}(M_{C})=\frac{1}{2C}\sum_{i=1}^{C}\frac{\delta_{ii}}{n_{i}}-\frac{1}{2C}\sum_{i=1}^{C}\frac{\bar{k}_{i}-\delta_{ii}}{n-n_{i}}.
\end{equation}
Now for simplicity without loss of generality if we assume that $n_{1} \leq n_{2} \cdots \leq n_{C}$, then for any $i > j, \forall i, j \in \mathcal{C}$, from (\ref{aucovaCond21}) we can conclude that, increase in $\bar{k}_{i}-\delta_{ii}$ (i.e. the misclassification to other classes) will have larger effect on the value of $\rho_{a}(M'_{C})$ than $\bar{k}_{j}-\delta_{jj}$. In other words, the cost of misclassifications is higher for the majority class. Hence the value of $\rho_{a}(M_{C})$ will be minimum when all the points from classes other than $C$ are wrongly predicted as class $C$, while the points from class $C$ are misclassified as $C-1$. Hence, from (\ref{aucova}) we get,
\begin{equation}
\label{aucovaCond22}
    \mathcal{L}_{\mathbb{M}_{C}}(\rho_{a})=\frac{1}{2C}\Big(C-1-\frac{n_{C}}{n-n_{C-1}} \Big).
\end{equation}
If the classifier correctly classifies all the test points then $k_{i}-m_{ii}=0$, while  $m_{ii}/n_{i}=1,  \forall i \in \mathcal{C}$. Plugging these values in (\ref{aucova}) gives $\mathcal{U}_{\mathbb{M}_{C}}(\rho_{a})=1$, thus finishing the proof.
\end{proof}

We can now state the following theorem which investigates the behavior of different indices under the effect of varying RRT and number of classes.
\begin{theorem}
\label{theoMultiCon1}
Given a $C$-class classification problem:
\begin{enumerate}
    \item The index $\gamma_{C}$ satisfies both of Condition \ref{cond1} and \ref{cond2}.
    \item The index $\alpha$ satisfies both of Condition \ref{cond1} and \ref{cond2}.
    \item The index $\rho_{o}$ satisfies Condition \ref{cond1} but not Condition \ref{cond2}.
    \item The index $\rho_{a}$ fails to satisfy both of Condition \ref{cond1} and \ref{cond2}.
    \item The index $\kappa_{a}$ satisfies Condition \ref{cond2} but not Condition \ref{cond1}.
\end{enumerate}
\end{theorem}
\begin{proof}
Let us consider two $C$-class confusion matrices $M_{C}$, and $M'_{C}$. If we define $M_{C}$ as per Definition \ref{confmat}, then we can construct a new confusion matrix $M'_{C}$ by multiplying all the elements in the $i^{th}$ row by a $b_{i}$ where, $b_{i} \in \mathbb{R}^{+}$, $b_{i}m_{ij} \in \mathbb{Z}^{+}$, and $b_{i} \neq b_{j}; \forall i, j, \in \mathcal{C}$, such that $M_{C} \sim M'_{C}$. 

1) Using (\ref{gmm}) we find the value of $\mathcal{V}_{M'_{C}}(\gamma_{C})$ as follows:
\begin{equation*}
    \mathcal{V}_{M'_{C}}(\gamma_{C})=\Bigg(\prod_{i=1}^{C}\frac{b_{i}m_{ii}}{b_{i}n_{i}} \Bigg)^{\frac{1}{C}}=\Bigg(\prod_{i=1}^{C}\frac{m_{ii}}{n_{i}} \Bigg)^{\frac{1}{C}}=\mathcal{V}_{M_{C}}(\gamma_{C}).
\end{equation*}
Hence, it is proved that $\gamma_{C}$ satisfies Condition \ref{cond1}.

Given a $C$-class confusion matrix $M_{C}$, the $\gamma_{C}(M_{C})$ is only dependent on the values of $m_{ii}$, and $n_{i}$, as can be inferred from its definition in (\ref{gmm}). Now, from Definition \ref{confmat} we know that the values of $n_{i}>0$, and $m_{ii} \geq 0$ (non-zero positive when at least one point from the class is correctly classified, 0 otherwise) for all $i \in \mathcal{C}$. Thus, from (\ref{gmm}), it is evident that $\gamma_{C}(M_{C})\geq 0$ (non-zero only when $m_{ii}>0; \forall i \in \mathcal{C}$), which implies that $\mathcal{L}_{\mathbb{M}_{C}}(\gamma_{C})=0$. 

Similarly, from the fact that $m_{ii}<=n_{i}; \forall i \in \mathcal{C}$, as $n_{i}=\sum_{j=1}^{C}m_{ij}$, we can conclude that $0 \leq m_{ii}/n_{i} \leq 1$. Therefore, from (\ref{gmm}) the value of $\mathcal{U}_{\mathbb{M}_{C}}(\gamma_{C})$ can found to be 1. Now, given the family of $C+1$-class confusion matrices, by the similar argument it can be shown that $\mathcal{L}_{\mathbb{M}_{C+1}}(\gamma_{C})=0$, and $\mathcal{U}_{\mathbb{M}_{C+1}}(\gamma_{C})=1$, which satisfies the Condition \ref{cond2}. This completes the first part of the theorem.

2) We take a $C$-class confusion matrix $M_{C}$, and construct $M'_{C}$, such that they belong to the same equivalence class. We then find the value of $\mathcal{V}_{M'_{C}}(\alpha)$ as per (\ref{gmm}):
\begin{equation*}
    \mathcal{V}_{M'_{C}}(\alpha)=\frac{1}{C}\sum_{i=1}^{C}\frac{b_{i}m_{ii}}{b_{i}n_{i}} =\frac{1}{C}\sum_{i=1}^{C} \frac{m_{ii}}{n_{i}}=\mathcal{V}_{M_{C}}(\alpha).
\end{equation*}
Therefore, $\alpha$ satisfies Condition \ref{cond1}.

It is evident from Lemma \ref{supp:lem1} that $\mathcal{L}_{\mathbb{M}_{C}}(\alpha)=0$, $\mathcal{U}_{\mathbb{M}_{C}}(\alpha)=1$. By the same logic it can be claimed that $\mathcal{L}_{\mathbb{M}_{C+1}}(\alpha)=0$, $\mathcal{U}_{\mathbb{M}_{C+1}}(\alpha)=1$. Therefore, the lower and upper bound of $\alpha$ does not change with the increase in the number of classes, proving the second part of the theorem.

3) We start by a $C$-class confusion matrix $M_{C}$, and construct $M'_{C}$, ensuring that $M_{C} \sim M'_{C}$. Now to confirm if $\rho_{o}$ satisfies Condition \ref{cond1}, we find $\mathcal{V}_{M'_{C}}(\rho_{o})$, using (\ref{aucovo}) as follows:
\begin{align*}
    \mathcal{V}_{M'_{C}}(\rho_{o})&=\frac{1}{2C}\sum_{i=1}^{C}\Bigg(1+\frac{b_{i}m_{ii}}{b_{i}n_{i}}-\sum_{\substack{j=1 \\ j \neq i}}^{C} \frac{b_{j}m_{ji}}{(C-1)b_{j}n_{j}}\Bigg), \nonumber \\
    =&\frac{1}{2C}\sum_{i=1}^{C}\Bigg(1+\frac{m_{ii}}{n_{i}}-\sum_{\substack{j=1 \\ j \neq i}}^{C} \frac{m_{ji}}{(C-1)n_{j}}\Bigg)=\mathcal{V}_{M_{C}}(\rho_{o}). 
\end{align*} Thus we show the invariance of $\rho_{o}$ under Condition \ref{cond1}.

We first start with a $M_{C} \in \mathbb{M}_{C}$, then following from Lemma \ref{lem1}, we get:
\begin{equation*}
    \mathcal{L}_{\mathbb{M}_{C}}(\rho_{o})=\frac{C-2}{2(C-1)} \;
    \text{and } \; \mathcal{U}_{\mathbb{M}_{C}}(\rho_{o})=1.
\end{equation*} 
Approaching similarly for a $C+1$-class confusion matrix, we see that:
\begin{equation}
\label{aucovoCond26}
    \mathcal{L}_{\mathbb{M}_{C+1}}(\rho_{o})=\frac{C-1}{2C} \neq \mathcal{L}_{\mathbb{M}_{C}}(\rho_{o}),
\end{equation}
\begin{equation}
\label{aucovoCond27}
    \mathcal{U}_{\mathbb{M}_{C+1}}(\rho_{o})=1=\mathcal{U}_{\mathbb{M}_{C}}(\rho_{o}).
\end{equation}
Therefore, from (\ref{aucovoCond26}), and (\ref{aucovoCond27}), we conclude that the lower bound of $\rho_{o}$ is dependent on the number of classes while the upper bound is remained at 1, violating Condition \ref{cond2} and completing the third part of the theorem.

4) Similar to the previous approaches given a $C$-class confusion matrix $M_{C}$, we construct $M'_{C}$, and express $\mathcal{V}_{M'_{C}}(\rho_{a})$ as follows:
\begin{align*}
    \mathcal{V}_{M'_{C}}(\rho_{a})&=\frac{1}{2C}\sum_{i=1}^{C}\left(1+\frac{b_{i}m_{ii}}{b_{i}n_{i}}-\frac{\sum_{j=1}^{C}b_{j}m_{ji}-b_{i}m_{ii}}{\sum_{j=1}^{C}b_{j}n_{j}-b_{i}n_{i}}\right), \nonumber \\
    =\frac{1}{2C}\sum_{i=1}^{C}&\left(1+\frac{m_{ii}}{n_{i}}-\frac{\sum_{j=1}^{C}b_{j}m_{ji}-b_{i}m_{ii}}{\sum_{j=1}^{C}b_{j}n_{j}-b_{i}n_{i}}\right) \neq \mathcal{V}_{M_{C}}(\rho_{a}).
\end{align*}
Hence, $\rho_{a}$ do not satisfy Condition \ref{cond1}.

It is evident from Lemma \ref{lem2} that for a $C+1$-class problem $\mathcal{U}_{\mathbb{M}_{C+1}}(\rho_{a})=\mathcal{U}_{\mathbb{M}_{C}}(\rho_{a})=1$. Moreover, similar to (\ref{aucovaCond22}), we can calculate:
\begin{equation}
\label{aucovaCond23}
    \mathcal{L}_{\mathbb{M}_{C+1}}(\rho_{a})=\frac{1}{2C+2}\Big(C-\frac{n_{C+1}}{n-n_{C}} \Big).
\end{equation}
From, (\ref{aucovaCond22}) and (\ref{aucovaCond23}) we can show $\mathcal{L}_{\mathbb{M}_{C}}(\rho_{a})\neq \mathcal{L}_{\mathbb{M}_{C+1}}(\rho_{a})$, indicating that $\rho_{a}$ does not satisfy Condition \ref{cond2}, which completes the fourth part of the theorem.

5) As previous we  take a $C$-class confusion matrix $M_{C}$, and construct $M'_{C}$ satisfying the equivalence relation. Let us now find $\mathcal{V}_{M'_{C}}(\kappa_{a})$ by (\ref{rpm}),
\begin{align*}
\label{rpmCond11}
    \mathcal{V}_{M'_{C}}(\kappa_{a})&=\frac{1}{2C}\sum_{i=1}^{C}\left(\frac{b_{i}m_{ii}}{\sum_{j=1}^{C}b_{j}m_{ji}}+\frac{b_{i}m_{ii}}{b_{i}n_{i}}\right), \nonumber \\
    &=\frac{1}{2C}\sum_{i=1}^{C}\left(\frac{b_{i}m_{ii}}{\sum_{j=1}^{C}b_{j}m_{ji}}+\frac{m_{ii}}{n_{i}}\right)\neq \mathcal{V}_{M_{C}}(\kappa_{a}).
\end{align*}
Therefore, we conclude that $\kappa_{a}$ violates Condition \ref{cond1}.

We know from Definition \ref{confmat}, if $m_{ii}$ becomes $n_{i}$ for all $i \in \mathcal{C}$, i.e. when all the points in the test set are correctly classified in their respective classes, then $k_{i}=m_{ii}; \forall i \in \mathcal{C}$. On the other hand if all the test points are misclassified then $m_{ii}=0$ for all $i \in \mathcal{C}$. Consequently, $0 \leq m_{ii}/k_{i} \leq 1$, and $0 \leq m_{ii}/n_{i} \leq 1$, where both reaches the lower bound of 0 when $m_{ii}=0$ (at the worst performance of the classifier), and the upper bound 1 when $m_{ii}=n_{i}$ (i.e. the classifier has achieved the best performance). Following this observation we can calculate $\mathcal{L}_{\mathbb{M}_{C}}(\kappa_{a})=\frac{1}{2}(0+0)=0$ and $\mathcal{U}_{\mathbb{M}_{C}}(\kappa_{a})=\frac{1}{2}(1+1)=1$. We can similarly find $\mathcal{L}_{\mathbb{M}_{C+1}}(\kappa_{a})$ and $\mathcal{U}_{\mathbb{M}_{C+1}}(\kappa_{a})$, which will be equal to their respective values for the set of $C$-class confusion matrices. Hence, $\kappa_{a}$ satisfies Condition \ref{cond2}.
\end{proof}

If we consider the case of AUROC-OVO then it is evident from Lemma \ref{lem1}, that the lower limit of $\rho_{o}$ gradually increases with the number of classes thus becomes affected by the Type 2 distortion. A solution to mitigate this problem is to apply a normalization to $\rho_{o}$, such that its lower bound can be made independent of $C$. This can be done by first subtracting the bias from $\rho_{o}$ and then dividing the result by the coefficient (both terms are dependent on the choice of $C$) found in (\ref{aucovoCond21}), which necessarily reduces the index to ACSA.

In a similar fashion we can discuss the nature of AUROC-OVA as well. As per Lemma \ref{lem2} the lower bound of the $\rho_{a}$ index is dependent on the number of classes as well as on the number of test points from the top two majority classes. Therefore, normalizing such an index will require to make certain assumptions on the representatives of the majority classes in the test set. In a special situation where the test set only contains $\frac{n}{2}$ points each from the top two majority classes, then $\mathcal{L}_{\mathbb{M}_{C}}(\rho_{a})$ reduces down to $\lambda_{C}=\frac{C-2}{2C}$, which is a weak lower limit for $\rho_{a}$. We call the normalized AUROC-OVA, as nAUROC-OVA, and calculate it by first subtracting the reduced lower limit and then dividing the result by the difference between the reduced lower limit and unity. In other words nAUROC-OVA can be expressed as $\frac{\rho_{a}(M_{C})-\lambda_{C}}{1-\lambda_{C}}$.

The reason for violating Condition \ref{cond1} by AURPC-OVA is the direct consideration of $k_{i}$ (which involve the true as well as false predictions in the $i^{th}$ class) in the precision counterpart. Therefore, we propose a modified AURPC-OVA such that while calculating the precision the $m_{ji}$ values are properly scaled by their corresponding $n_{j}$s, for all $j, i \in \mathcal{C}$. We describe the modified AURPC-OVA called as mAURPC-OVA, in the following Definition \ref{mrpm}.
\begin{definition}
\label{mrpm}
For a $C$-class confusion matrix $M_{C}$ the mAURPC-OVA index, denoted by $\hat{\kappa_{a}}$ is defined as:
\begin{equation}
    \hat{\kappa_{a}}(M_{C})=\frac{1}{2C}\sum_{i=1}^{C}\left(\frac{m_{ii}/n_{i}}{\sum_{j=1}^{C}m_{ji}/n_{j}}+\frac{m_{ii}}{n_{i}}\right).
\end{equation}
\end{definition}
We now proceed to confirm Condition \ref{cond1}, and \ref{cond2} for mAURPC-OVA, in the following theorem.
\begin{theorem}
\label{mrpmCond1}
The index $\hat{\kappa_{a}}$ satisfies both of the Conditions.
\end{theorem}
\begin{proof}
Proceeding in a manner similar to the one taken for $\kappa_{a}$ in Theorem \ref{theoMultiCon1}, if we consider the two $C$-class confusion matrices $M_{C}$ and $M'_{C}$, then $\mathcal{V}_{M'_{C}}(\hat{\kappa_{a}})$ can be expressed as follows:
\begin{align*}
    \mathcal{V}_{M'_{C}}(\hat{\kappa_{a}})&=\frac{1}{2C}\sum_{i=1}^{C}\left(\frac{b_{i}m_{ii}/b_{i}n_{i}}{\sum_{j=1}^{C}b_{j}m_{ji}/b_{j}n_{j}}+\frac{b_{i}m_{ii}}{b_{i}n_{i}}\right), \nonumber \\
    &=\frac{1}{2C}\sum_{i=1}^{C}\left(\frac{m_{ii}/n_{i}}{\sum_{j=1}^{C}m_{ji}/n_{j}}+\frac{m_{ii}}{n_{i}}\right)= \mathcal{V}_{M_{C}}(\hat{\kappa_{a}}),
\end{align*}
which indicates that $\hat{\kappa_{a}}$, satisfies Condition \ref{cond1}.

Similar to Theorem \ref{theoMultiCon1}, we can see that $0 \leq m_{ii}/n_{i} \leq 1$, and $0 \leq \sum_{j=1}^{C}m_{ji}/n_{j} \leq 1$, for all $i, j, \in \mathcal{C}$. Both of these terms reach their corresponding lower bound when classifiers performs the worst and upper bound at the accurate classification as previously described in the fifth part of Theorem \ref{theoMultiCon1}. Therefore, following Theorem \ref{theoMultiCon1}, we can conclude that both of  $\mathcal{L}_{\mathbb{M}_{C}}(\hat{\kappa_{a}})=0=\mathcal{L}_{\mathbb{M}_{C+1}}(\hat{\kappa_{a}})$ and $\mathcal{U}_{\mathbb{M}_{C}}(\hat{\kappa_{a}})=\mathcal{U}_{\mathbb{M}_{C+1}}(\hat{\kappa_{a}})$, hold implying that Condition \ref{cond2} is satisfied by $\hat{\kappa_{a}}$.
\end{proof}

From Theorem \ref{theoMultiCon1} and \ref{mrpmCond1} we can conclude that among all only GMean, ACSA, and AURPC-OVA satisfy both of Condition \ref{cond1} and \ref{cond2} and thus can be applicable to evaluate multi-class imbalanced classifiers in presence of varying RRT or number of classes. However, the question about the quality of information provided by these indices under extremely poor classification performance on a single class is still required to be answered. Therefore, we proceed to the following Theorem \ref{cond3check} which evaluates the indices under the light of Condition \ref{cond3}. 

\begin{theorem}
\label{cond3check}
Among the three indices which are immune to the two types of distortions, except $\gamma_{C}$ both of $\alpha$, and $\hat{\kappa_{a}}$ also satisfy Condition \ref{cond3}.
\end{theorem}

\begin{proof}
To prove that $\gamma_{C}$ fails to satisfy the third condition we start by finding $\mathcal{W}_{\mathcal{M}_{C}(i)}(\gamma_{C})$, which by (\ref{gmm}) can be expressed as follows:
\begin{equation}
\label{gmCond31}
    \mathcal{W}_{\mathcal{M}_{C}(i)}(\gamma_{C})=\lim_{\epsilon \to 0^{+}}\Bigg(\epsilon \prod_{j=1, j \neq i}^{C}(1-\epsilon)\Bigg)^{\frac{1}{C}}=\lim_{\epsilon \to 0^+}\epsilon^{\frac{1}{C}}(1-\epsilon)^{\frac{C-1}{C}}=0. 
\end{equation}
From (\ref{gmCond31}) and Theorem \ref{theoMultiCon1}, we can see that for $\gamma_{C}$ index $\mathcal{W}_{\mathcal{M}_{C}(i)}(\gamma_{C})=\mathcal{L}_{\mathbb{M}_{C}}(\gamma_{C})$, thus violating Condition \ref{cond3}. 

We begin by calculating $\mathcal{W}_{\mathcal{M}_{C}(i)}(\alpha)$, which according to (\ref{gmm}) is as follows:
\begin{align}
\label{acsaCond31}
    \mathcal{W}_{\mathcal{M}_{C}(i)}(\alpha)&=\lim_{\epsilon \to 0^+}\frac{1}{C}\Big(\epsilon + \sum_{j=1, j \neq i}^{C}(1-\epsilon)\Big), \nonumber \\
    &=\lim_{\epsilon \to 0^+}\frac{\epsilon+(C-1)(1-\epsilon)}{C}=\frac{C-1}{C}. 
\end{align}
From (\ref{acsaCond31}) and Theorem \ref{theoMultiCon1}, it is evident that $\alpha$ index $\mathcal{W}_{\mathcal{M}_{C}(i)}(\alpha) > \mathcal{L}_{\mathbb{M}_{C}}(\alpha)$, therefore, $\alpha$ satisfies Condition \ref{cond3}. 

As per (\ref{mrpm}) the value of $\mathcal{W}_{\mathcal{M}_{C}(i)}(\hat{\kappa_{a}})$ can be calculate as:
\begin{align}
\label{mrpmCond31}
\begin{split}
    \mathcal{W}_{\mathcal{M}_{C}(i)}(\hat{\kappa_{a}})&=\lim_{\epsilon \to 0^+} \frac{1}{2C}\sum_{\substack{j=1 \\ j \neq i}}^{C}\Bigg(1-\epsilon+ \frac{1-\epsilon}{1-\epsilon+\frac{k_{i}-m_{jj}}{n-n_{j}}}\Bigg) \\
    &+ \lim_{\epsilon \to 0^+}\frac{1}{2C}\Bigg(\epsilon+\frac{\epsilon}{\epsilon+ \frac{k_{i}-m_{ii}}{n-n_{i}}}\Bigg),
\end{split} \nonumber \\
\begin{split}
    \Rightarrow \mathcal{W}_{\mathcal{M}_{C}(i)}(\hat{\kappa_{a}})&>\lim_{\epsilon \to 0^+} \frac{1}{2C}\sum_{\substack{j=1 \\ j \neq i}}^{C}\Bigg(1-\epsilon+ \frac{1-\epsilon}{1-\epsilon+\frac{n-n_{j}+n_{j}\epsilon}{n-n_{j}}}\Bigg) \\
    &+\lim_{\epsilon \to 0^+}\frac{1}{2C} \Bigg( \epsilon+\frac{\epsilon}{\epsilon+ \frac{n-n_{i}\epsilon }{n-n_{i}}} \Bigg)
\end{split} \nonumber \\
    \Rightarrow \mathcal{W}_{\mathcal{M}_{C}(i)}(\hat{\kappa_{a}})&> \frac{C-1}{2C}(1+ \frac{1}{2})+0=\frac{3(C-1)}{4C}.
\end{align}
From (\ref{mrpmCond31}) and Theorem \ref{mrpmCond1}, it is evident that $\mathcal{W}_{\mathcal{M}_{C}(i)}(\hat{\kappa_{a}})>\mathcal{L}_{\mathbb{M}_{C}}(\hat{\kappa_{a}})$, confirming that the index $\hat{\kappa_{a}}$ satisfies Condition \ref{cond3}, which completes the proof.
\end{proof}

\section{Experiments and results} \label{res}
This section first provides a brief description of the used dataset, followed by details of the experiment protocol and finally illustrates the different results alongside appropriate discussion.

\subsection{Description of datasets}
We have used a subset of the widely popular ImageNet \cite{imagenet} classification dataset for all our experiments. The ImageNet dataset provides a vast collection of natural images categorized into a large number of structured classes (1000 leaf classes alongside 860 higher level concepts following a predefined tree). For our experiments we have taken a subset of the ImageNet training set by sampling images from 12 higher level classes (formed by combining 1-5 leaf concepts and containing a total of 1300-6500 data instances). Since our chosen classifiers are only applicable to real valued data, given the images, we need to extract quality features. Therefore, for the purpose of feature extraction we have used the state-of-the-art Inception V3 \cite{szegedy2016}, an end-to-end deep neural network, which learns the map from the image space to a set of classes through a 2048-dimensional real values distributed representation space. The Inception V3 used by us is a standard implementation pre-trained on the complete ImageNet training set, publicly available from Keras deep learning API at \url{https://keras.io/applications}. Thus, in our case the selected subset of images can be mapped to useful feature vectors by a simple forward pass through the pre-trained network. We have then created 12 two-class classification problems, having $IR$ between 5 and 40 for experimentally validating the effect of Type 1 distortion over the various indices. Moreover, we have formed a total of 10 multi-class imbalanced datasets (4 sets for 3-class, while 3 sets each for the 5-class and 10-class classification problems) having $IR$ between 20-30 for the purpose of empirically evaluating the effect of Type 2 distortion. A detailed description of the datasets used in our experiments can be found in Section 2 of the supplementary document.

\subsection{Experiment protocol}
We perform two sets of experiments respectively over two-class and multi-class datasets to inspect the behavior of different indices in light of Condition \ref{cond1} and Condition \ref{cond2}. We conduct our experiments using four state-of-the-art, classifiers of diverse nature, all specifically tailored for handling class-imbalance, namely Dual-LexiBoost with $k$-Nearest Neighbor as the base classifier \cite{datta2019boosting}, Near Bayesian SVM (NBSVM) \cite{datta2015}, RUSBoost \cite{seiffert2010, Japkowicz2000} with decision trees \cite{breiman2017} as the base classifier, and MLP \cite{rumelhart1986} combined with SMOTE \cite{chawla2002}. The parameter settings of these methods can be found in Section 3 of the supplementary material.  

For both experiments, the classifier is first trained with the training set and then tested by multiple test sets having different RRT values. This is done in an attempt to mimic the two primary causes of the first type of distortion as described in Section \ref{sec:motiv}. In case of two-class datasets the RRT is varied between 1 (balanced), 0.5 (more number of minority points are taken compared to majority), half of the original IR (reduced effect of imbalance), the original IR of the training set, and twice of the original IR. Similarly in case of multi-class datasets, we have used 5 different test sets with varying RRT (The first is balanced, in the second the IR between the classes are reversed, in the third the original IR between the minority class and all others are halved, the fourth maintains the original IR, while the last doubles the test points from all classes except the minority). Such an experimental setting helps us to understand the effect of the varying number of test points from different classes on the values of the indices. Additionally, the experiments on multi-class datasets provide us with a way to inspect the effect of increasing $C$ on the indices. Average results over 5 independent runs of 5-fold stratified cross validation (which also aids in parameter tuning for the classifiers) are reported to ensure the reliability of our findings.

\begin{figure}
    \centering
    \includegraphics[width=6cm]{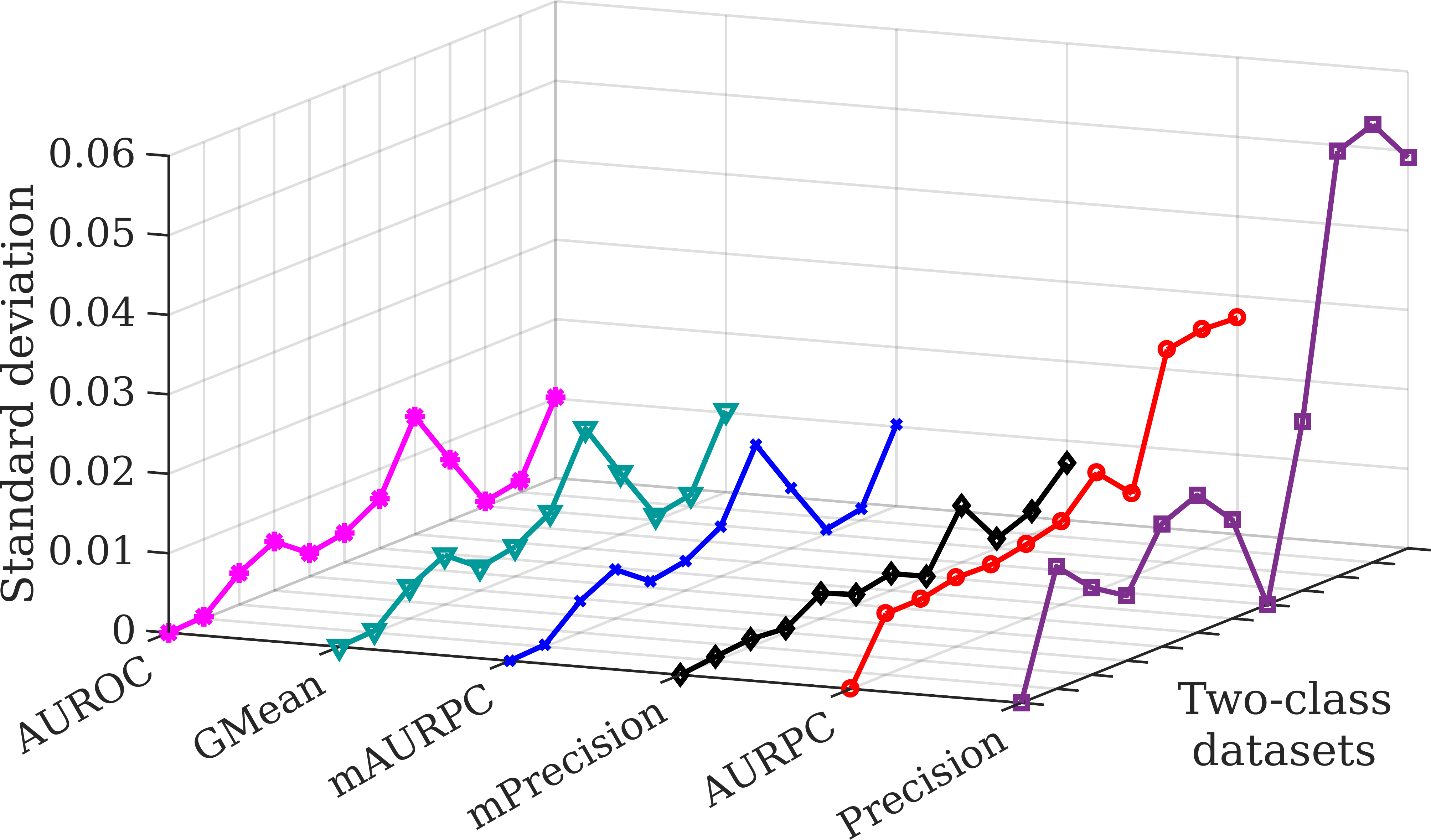}
    \caption{Effect of RRT on different indices over two-class datasets}
    \label{twoClsStd}
\end{figure}

\subsection{Validating the two-class classification performance evaluation indices in light of Condition \ref{cond1}}
For each of the dataset, we have found the standard deviation of the mean performance in terms of an index over the five test sets and four classifiers. We plot the findings in Figure \ref{twoClsStd}, which shows that the Precision index achieves the highest variability over the test sets for a given training set. However, the low standard deviation of GMean and AUROC suggests that the classifiers retain an almost similar performance over the various test sets. Therefore, the high standard deviation of Precision must be due to the changes in the actual numbers of the respective test points from the two classes, which vary significantly due to the diverse choice of RRT. These observations reflect the theoretical analysis which shows Precision to be sensitive over RRT even when the class-specific classification performance is retained, thus failing to satisfy Condition \ref{cond1}. Due to having Precision as a component AURPC also suffers from the same issue, though the additional consideration of Recall helps to mitigate the effect of altering RRT to some extent. Interestingly, mPrecision and mAURPC closely follow the GMean and AUROC indices indicating their immunity against the effect of RRT. 

\begin{figure}[!ht]
    \centering
    \subfloat[\label{multiClsStd}]{\includegraphics[width=0.47\textwidth]{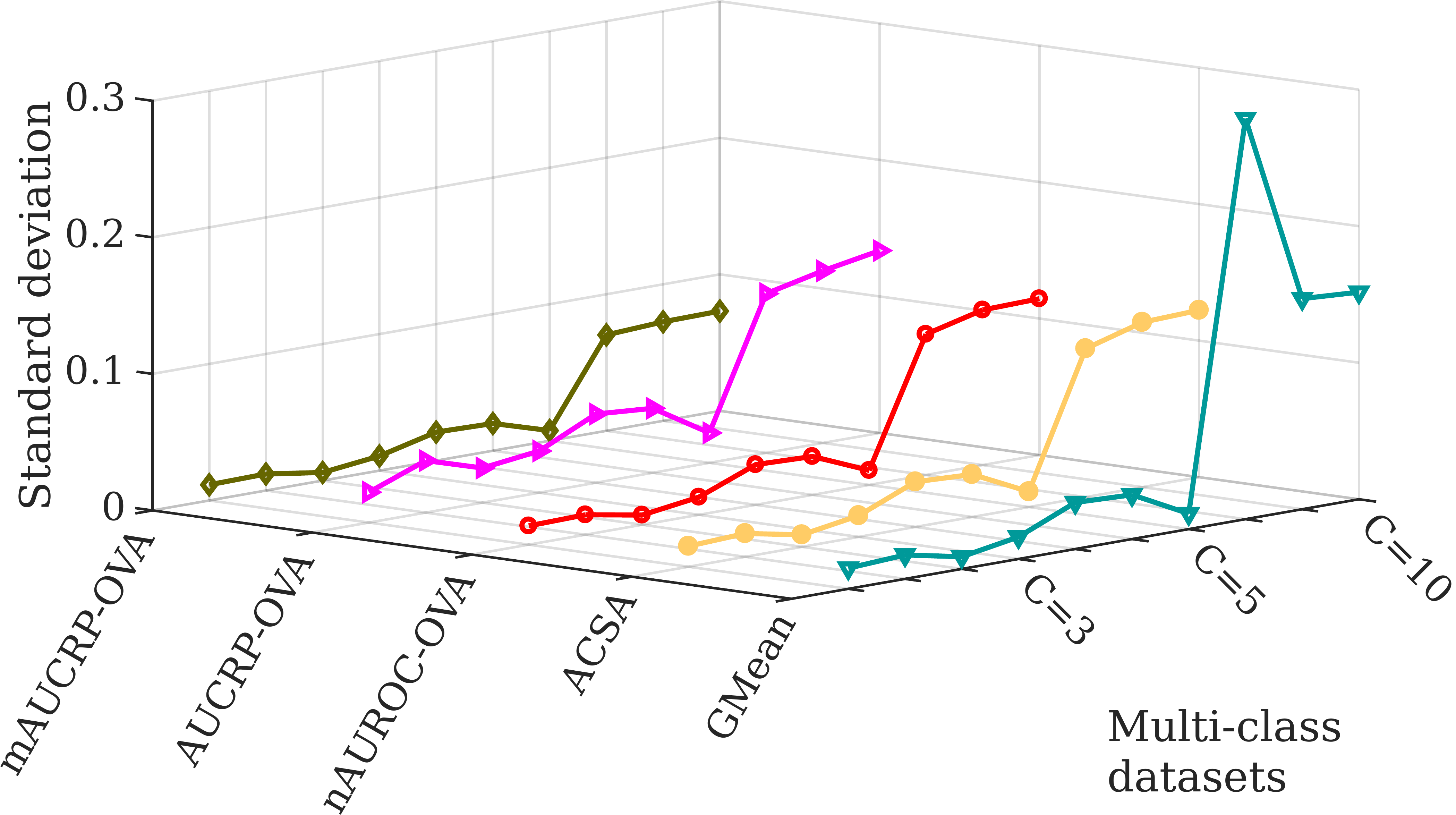}} 
    \hspace{3mm}
    \subfloat[\label{stability}]{\includegraphics[width=0.47\textwidth]{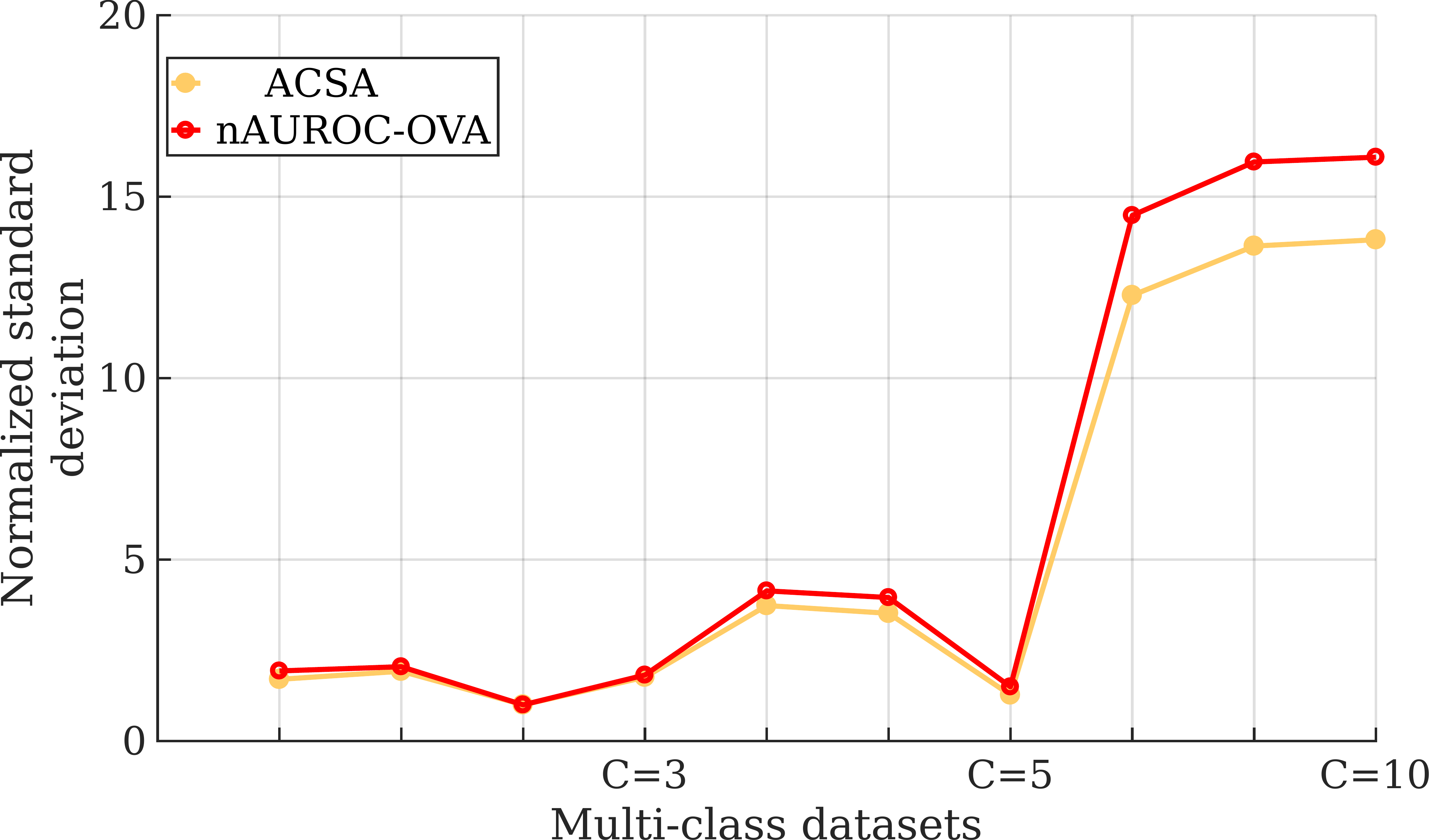}}
    \caption{Analysis of index behavior over multi-class datasets under Condition \ref{cond1}. (a) Effect of RRT on different indices over multi-class datasets. (b) Stability of ACSA compared to nAUROC-OVA.}
    \label{fig:multiClassCond2}
\end{figure}

\subsection{Validating the multi-class classification performance evaluation indices in light of Condition \ref{cond1}}
We use an approach similar to the two-class case for validating Condition 1 for the multi-class performance evaluation indices. However, the indices which are susceptible to Condition 2 are expected to have an smaller range with increasing value of $C$, and may result into a lower standard deviation over the test sets for high number of classes. Thus, comparing these indices with those indices satisfying Condition 2 may lead to a bias against the later and will not help to reach a conclusive remark. Hence, in Figure \ref{multiClsStd}, we only compare the standard deviations of indices satisfying Condition \ref{cond2}, viz. GMean, ACSA, nAUROC-OVA, AURPC-OVA, and mAURPC-OVA, over the various test sets for each of the datasets. A close inspection reveals that the minimum variability (especially improving from AURPC-OVA) is achieved by mAURPC-OVA establishing it as the better choice among the five contenders. Interestingly, ACSA has shown slightly higher variability compared to nAUROC-OVA, which is unlikely as the later violates Condition \ref{cond1}. This leads us to investigate further, by normalizing the standard deviation of the ACSA and nAUROC-OVA indices for each of the datasets by the respective minimum standard deviation achieved over all the multi-class datasets. This kind of normalized standard deviation can be considered as a measure of stability as it quantifies the variability of an index from its best stable performance (a lower value signifies that the index can equivalently evaluate similar performing classifiers). We plot the results in Figure \ref{stability}, which shows the normalized standard deviation to be slightly greater for nAUROC-OVA than that of ACSA. This indicates nAUROC-OVA to be less stable compared to ACSA, and the lower standard deviation of the former in Figure \ref{multiClsStd} may be due to the fact that AUROC-OVA is normalized using a weak lower bound.

\subsection{The effect of the number of classes (Condition \ref{cond2}) over the different indices}
\begin{figure}[!ht]
    \centering
    \includegraphics[width=6cm]{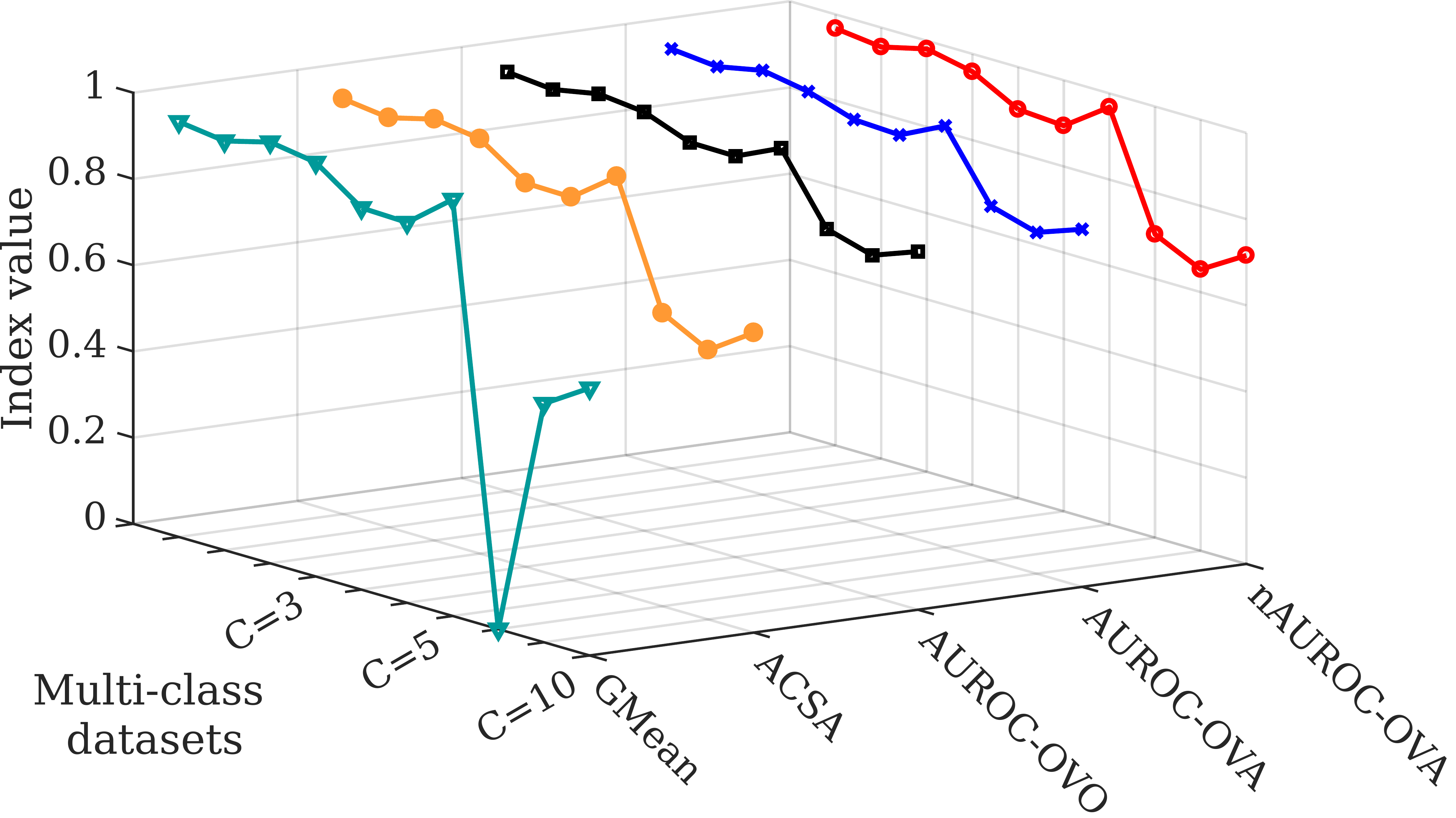}
    \caption{Effect of the number of classes on different indices over multi-class datasets}
    \label{classEffect}
\end{figure}

We consider AUROC-OVA, and AUROC-OVO for this experiment as they are seen to have a higher lower bound with increasing number of classes. Their respective normalized version, i.e. nAUROC-OVA and ACSA are also considered to establish the improvement achieved through normalization, alongside GMean as a reference. We plot the minimum value achieved by these indices for each of the datasets in Figure \ref{classEffect}. The results shows that on three and five class datasets the AUROC-OVA, and AUROC-OVO performs almost equivalently to their normalized counterparts. However, on the ten class datasets the minimum index value achieved by AUROC-OVA and AUROC-OVA are significantly higher than ACSA, and nAUROC-OVA. This validates the bias of AUROC-OVA and AUROC-OVO towards a higher value with increasing number of classes, and also demonstrated the ability of the respective normalized versions to counter this bias.

\subsection{The effect of Condition \ref{cond3} on multi-class indices}
From Figure \ref{classEffect} the minimum values of GMean are consistent with the other indices over the three, and five class datasets. However, for the ten-class datasets the index produced significantly lower values compared to the others. Moreover, on a ten-class dataset GMean produced its lowest possible value of 0, indicating the worst possible classification performance. However, the values of the other indices over the same dataset clearly indicate that the classifier managed to successfully classify many of the test points. Therefore, despite satisfying Conditions \ref{cond1} and \ref{cond2}, GMean fails to do the same for Condition \ref{cond3} as poor performance on a single class results in the loss of all information about the performance on every other class. 

\begin{figure}[!ht]
    \centering
    \subfloat[\label{fig:recoFig}]{\includegraphics[width=0.35\textwidth]{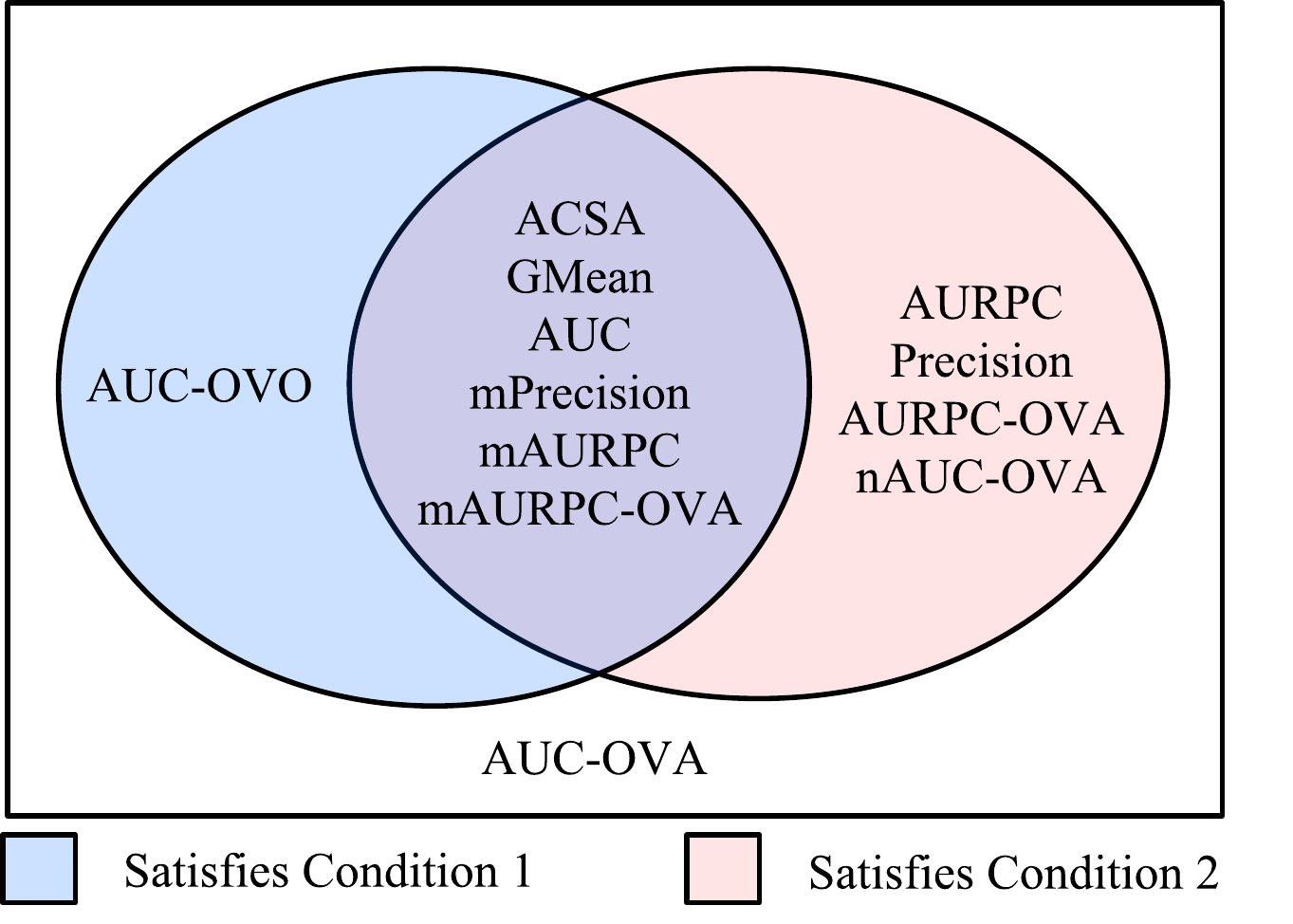}} 
    \hspace{3mm}
    \subfloat[\label{fig:recoFig2}]{\includegraphics[width=0.35\textwidth]{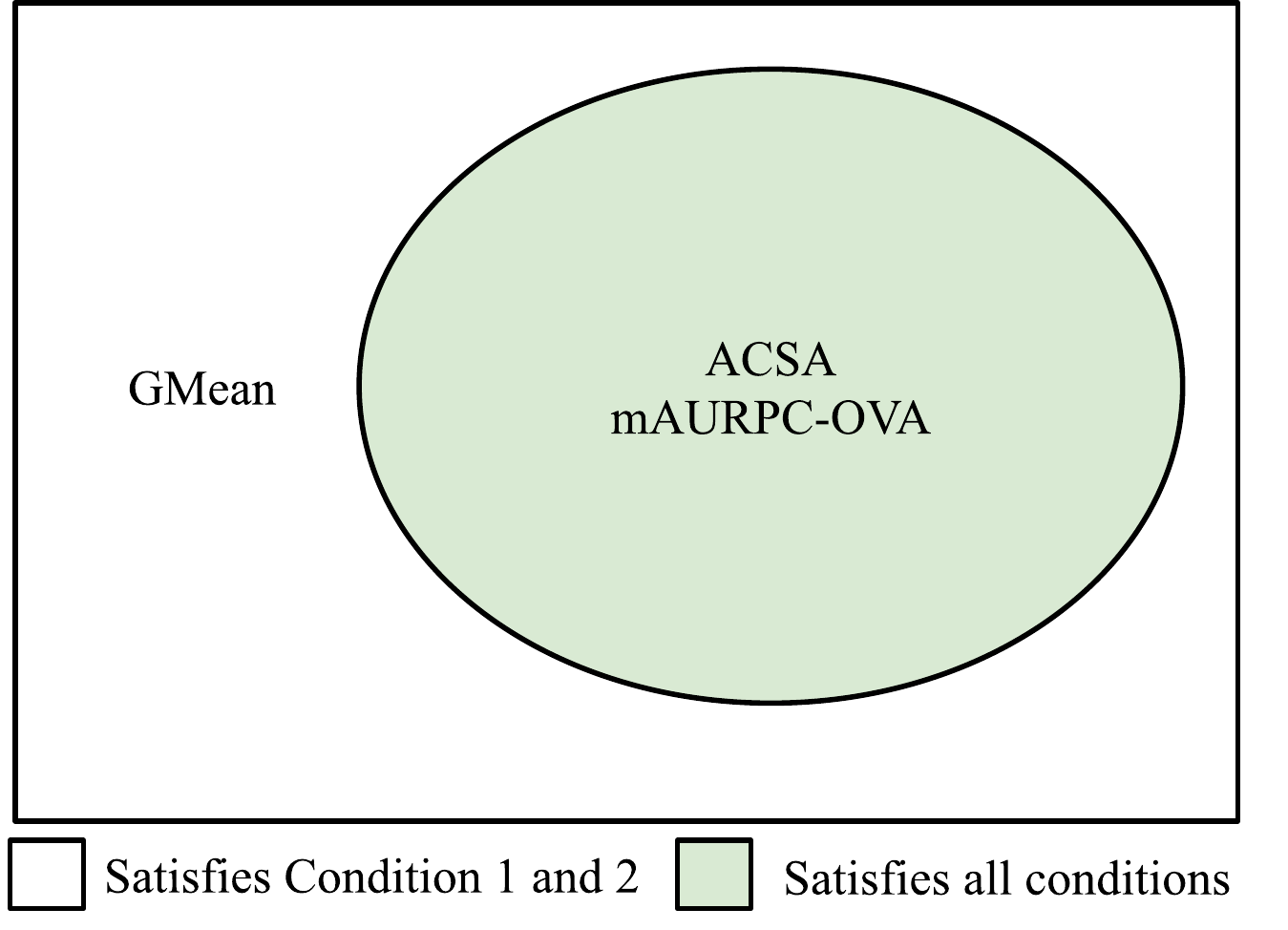}}
    \caption{A summary of the different conditions satisfied by each of the indices under concern. (a) Summary of findings documented by Theorem \ref{theo1}, \ref{theo2}, \ref{theoMultiCon1}, and \ref{mrpmCond1}, i.e. validation of indices under Condition \ref{cond1}, and \ref{cond2}. (b) Summary of findings in Theorem \ref{cond3check}, i.e. validation of the indices under the light of Condition \ref{cond3}, which satisfy the fundamental properties and applicable to multi-class classification problems.}
    \label{fig:reco}
\end{figure}

\section{Discussion on the applicability of indices} \label{discuss}
Based on the satisfaction of the two fundamental conditions the indices can be grouped as shown in Figure \ref{fig:recoFig}. Moreover, the multi-class indices which satisfy Condition \ref{cond1}, and \ref{cond2}, are further classified by Condition \ref{cond3} in Figure \ref{fig:recoFig2}. Therefore, using Figure \ref{fig:reco} we can proceed to recommend an appropriate choice of indices for different applications.

In case of two-class classification, all four of GMean, AUROC, mPrecision, and mAURPC satisfy Condition 1, thus any one of these can be a good index of choice. However, GMean is biased towards the accuracy of that class which is poorly classified compared to the other. In a two-class scenario, this property of GMean may prove useful as it will identify the high bias of a classifier towards a particular class. AUROC, on the other hand, accords equal weight to the performance in both classes. Therefore, we recommend GMean for general evaluation of the performance of a two-class classifier. 

Recall and Precision (consequently AURPC, mPrecision, and mAURPC) both depends on the choice of the positive class. Recall is focused on the classification performance over the minority class, thus can be used in applications where false positives do not lead to severe consequences. For example, we may consider the case of benign and malignant tumor classification in medical diagnostic systems, where wrongly classifying a sample from the minority class of malignant tumors may result in fatal outcome. On the other hand, Precision (and AURPC) can be effectively used when the application attempts to limit the number of false positives while the class priors do not significantly vary over time. One can think of the spam filtering problem where even though the non-spam mails are considerably high in numbers, labelling one of them as spam may lead to loss of important information. Evidently, mPrecision and mAURPC indices can act as the respective replacement of Precision and AURPC if the application under concern can cause Type 1 distortion. 

In case of multi-class classification, even though GMean satisfies both of Condition \ref{cond1} and \ref{cond2} it may still be biased in case of extremely poor performance over a single class, as indicated by its violation of Condition \ref{cond3}. GMeans can however still prove beneficial if the target is to achieve non-zero classification accuracy on each class. On the contrary, ACSA and mAURPC satisfy all three of the conditions, and thus any of the two can be an appropriate choice of index. Finally, despite their violation of Condition \ref{cond1}, nAUROC-OVA and AURPC-OVA can be used in those applications where the misclassification from different classes are associated with different costs. For example, in a multi-class medical diagnostic application a somewhat similar set of syndromes may correspond to different diseases of varying severity and rarity. 

\section{Conclusion and future works} \label{conclude}
In this article, we show that the common indices used for evaluating the performance of a classifier in presence of class imbalance may suffer from different forms of distortions, depending on the character of the data, especially the test set. We formally define two conditions that an index needs to satisfy to be resilient to such distortions. We present theoretical analyses detailing the traits of the indices in light of these conditions and propose necessary remedies as per need. We further define a third condition to evaluate the quality of the information provided by an index, especially under adverse conditions such as exceptionally poor accuracy over a single class. We also undertake empirical analysis to support our theoretical findings. Finally, we discuss on the applicability of different indices and make recommendations. 

A natural future extension of this work would be to investigate the behavior of indices which are used in imbalanced multi-label \cite{madjarov2012extensive} and multi-instance \cite{carbonneau2018multiple} classification problems. One may also consider validating the efficacy of the modified/normalised indices on class imbalanced problems where the two types of distortions are naturally occurring. For example, Type 1 distortion is quite inherent in image foreground and background classification \cite{lin2017focal}. This is because foreground usually spreads over less number of pixels compared to background resulting in class imbalance. Moreover, the the fraction of background to foreground in an image i.e. the RRT may significantly vary between images consequently causing distortion in performance indices which fail to satisfy Condition \ref{cond1}. Type 1 distortion is also possible during sentiment analysis from tweets, where not all sentiments may occur with equal frequency \cite{Zimbra2018}, while the prior probability of different sentiments may notably change over time. Thus, if a sentiment analyzing classifier is periodically tested after deployment for potential fine tuning and an index susceptible to Type 1 distortion is utilized for quantifying its performance, then the result may mislead the quality assessment. On the other hand, Type 2 distortion can occur in open set recognition or incremental learning problems where the number of classes may increase over time \cite{rudd2017extreme}, thus use of an index which satisfies Condition \ref{cond2} may be beneficial. 

\section*{Acknowledgement}
We gratefully acknowledge NVIDIA Corporation for donating us the Titan Xp GPU which was immensely useful in conducting this research.

\section*{References}
\bibliographystyle{elsarticle-num}
\bibliography{paper}

\end{document}